\documentclass[journal]{IEEEtran}
\IEEEoverridecommandlockouts
\usepackage{cite}
\usepackage{amsmath,amssymb,amsfonts}
\usepackage{algorithmic}
\usepackage{graphicx}
\usepackage{textcomp}
\usepackage[table]{xcolor}
\usepackage{amsmath}
\usepackage{tabularx}
\usepackage{verbatim}
\usepackage{siunitx}
\usepackage{array}
\usepackage[T1]{fontenc}
\usepackage{selinput}
\usepackage{soul}
\SelectInputMappings{%
  adieresis={ä},
}
\newcolumntype{P}[1]{>{\centering\arraybackslash}p{#1}}
\usepackage{placeins}
\usepackage[outline]{contour}
\definecolor{urban}{HTML}{00FFFF}
\definecolor{rangeland}{HTML}{FF00FF}

\def\BibTeX{{\rm B\kern-.05em{\sc i\kern-.025em b}\kern-.08em
    T\kern-.1667em\lower.7ex\hbox{E}\kern-.125emX}}
    
\begin{document}

\title{Improving land cover segmentation across satellites using domain adaptation\\
\thanks{This research was supported by Business Finland under Grant no. 1259/31/2018}
}

\author{Nadir Bengana,
        Janne Heikkilä
}

\maketitle

\begin{abstract}
Land use and land cover mapping are essential to various fields of study, including forestry, agriculture, and urban management. Using earth observation satellites both facilitate and accelerate the task. Lately, deep learning methods have proven to be excellent at automating the mapping via semantic image segmentation. However, because deep neural networks require large amounts of labeled data, it is not easy to exploit the full potential of satellite imagery. Additionally, the land cover tends to differ in appearance from one region to another; therefore, having labeled data from one location does not necessarily help in mapping others. Furthermore, satellite images come in various multispectral bands—the bands could range from RGB to over twelve bands.
In this paper, we aim at using domain adaptation to solve the aforementioned problems. We applied a well-performing domain adaptation approach on datasets we have built using RGB images from Sentinel-2, WorldView-2, and Pleiades-1 satellites with Corine Land Cover as ground-truth labels. We have also used the DeepGlobe land cover dataset.
Experiments show a significant improvement over results obtained without the use of domain adaptation. In some cases, an improvement of over 20\% MIoU. At times it even manages to correct errors in the ground-truth labels.
\end{abstract}


\section{Introduction}
\IEEEPARstart{O}{ver} the last few years, Remote Sensing (RS) data became easily obtainable thanks to the surge of open data from several Earth Observation (EO) satellites. Data such as Sentinel-2 from the European Copernicus program \cite{copernicus_open} and Landsat from the U.S. Geological Survey (USGS) \cite{usgs} are available for the public free of charge. These satellites provide high-resolution multispectral imagery (up to 10m), which facilitates the application of multiple methods in processing the data.

Land cover represents the (bio)physical cover on the earth's surface, while land use is the cover caused by the action done by humans, e.g., a wheat field is a part of land use while an ocean is strictly land cover. Land Use and Land Cover (LULC) mapping is one of the most crucial RS applications providing a way to monitor, for example, forests, agriculture, and oceans. The mapping can be done manually by looking through satellite images \cite{doi:10.1080/01431169208904145}. However, it is very costly and time-consuming. Additionally, a fine global land cover map does not exist. Corine Land Cover (CLC) \cite{corinepdf} is an example of what is already available with a pixel resolution of 100m/px; however, it covers Europe only, and it is updated only roughly once every six years. Moderate Resolution Imaging Spectroradiometer (MODIS) \cite{MODIS} provides a global land cover map that is updated annually, but the pixel resolution is 500m/px, which might be too coarse for many applications such as urban cover monitoring.

There exist several ways to perform LULC mapping automatically depending on the available data and on the desired accuracy. The simplest one is land cover classification, which labels patches depending on the majority land cover type. Pixel segmentation for land use and land cover is more challenging to perform than classification. Lately, classical Machine Learning (ML) tools have fallen out of favor regarding the task of semantic segmentation, including LULC segmentation. Otávio \textit{et al.} \cite{7301382} showed that Convolutional Neural Networks (CNN) vastly outperform the classical ML methods when it comes to land cover classification. In the land cover segmentation section of the DeepGlobe challenge \cite{DBLP:journals/corr/abs-1805-06561}, the leaderboards are completely dominated by Deep Neural Networks (DNN). In land cover segmentation, \cite{deepglobe.densenet}, \cite{DBLP:journals/corr/HuangLW16a}, and \cite{Kuo2018DeepAN} are leading examples where they rely on DNNs such as ResNet and DenseNet.

Deep Learning (DL) methods perform well, however, they require a huge amount of data to show their true potential. As mentioned previously, EO data is available for free, but fine Ground-Truth (GT) labels are rare. Although CLC has a resolution of 100m/px resolution, a finer version exists with 20m/px covering Finland provided by the Finnish environment institute \cite{syke}, and a 10m/px version covering Germany by Geodatenzentrum \cite{CLC_GR}. Although these versions cover a limited area and have a few inaccuracies, they can still be used to train a DNN for land cover segmentation.

Because land cover types are different depending on the location, having a DNN model that performs land cover segmentation for one area does not mean that it works for other areas. Additionally, images captured from various satellites are different due to the mismatch in capture time, pixel resolution, radiometric resolution, and other properties. All these variables result in a domain shift between a dataset acquired from a specific satellite covering a particular region and a second satellite covering either the same region or a different one. Therefore, to get consistent results, a model must be trained on each variation of the data, requiring massive labeled datasets from each satellite, which is unreasonable. 

Removing the domain shift between different datasets is called Domain Adaptation (DA). In the context of semantic segmentation, domain adaptation is used to segment images from a target dataset using a source dataset. There exist two types of domain adaptation approaches: supervised, where some or all of the target data is labeled, and unsupervised, where the target data is unlabeled. Domain adaptation has the potential to improve over results achieved with simple transfer learning.

This paper focuses on applying deep domain adaptation to segment RGB satellite data from different satellites and locations semantically.
Our contributions are:
\begin{enumerate}
\item To the best of our knowledge this is the first paper that applies domain adaptation to achieve global LULC mapping using RGB bands only.
\item Building RGB datasets from Sentinel-2, WorldView-2, and Pleiades-1 satellites using CLC as labels.
\item Customizing an existing domain adaptation method for satellite imagery.
\end{enumerate}
 
In Section II, we present some of the related background research. Section III  presents the materials used, including the satellite images and labels. Section IV shows the Methods used. Furthermore, in Section V, we display the experiments done and the results obtained from them. Finally, Section VI is the conclusion where we present our thoughts about the results obtained.

\section{related work}

\subsubsection{LULC segmentation}

In LULC, image segmentation is performed differently compared to other fields such as street-view images because satellite imagery comes in multispectral forms, including active sensing images like Synthetic-Aperture Radar (SAR) or LIDAR\footnote{LIDAR uses light in the form of a pulsed laser to measure ranges.}  

Using computer vision to map land use and land cover started soon after the launch of Landsat-1 in 1972, which is the first publicly available remote sensing satellite. Various manually tuned algorithms available for image classification and segmentation were applied to satellite imagery. Early on, methods such as histogram thresholding \cite{1455595} provided some acceptable results but suffered from problems caused by the variations in satellite images. Research on other methods, mainly statistical ones based on Maximum A Posteriori (MAP) had varying degrees of success. By the year 2000, the commercial software eCognition \cite{ecognition} gathered classification methods, edge detection, and segmentation into one solution. Neural network methods at this point were unfavorable due to their high computational complexity.

In the early 2000s, classical machine learning methods such as support vector machines and decision trees \cite{doi:10.1080/10106049.2013.768300} \cite{KHATAMI201689} were mainly applied. By the 2010s, Deep Neural Networks were starting to emerge but were still not used much in remote sensing imagery due to the lack of labeled data to train the required DNNs for the task. In particular, using deep learning for tackling the task of land cover segmentation has not been much addressed in the literature so far. 
As an example, the state of the art semantic segmentation DNN based methods were translated to satellite imagery. However, the results were not as good \cite{deepglobe.densenet} \cite{DBLP:journals/corr/HuangLW16a} \cite{Kuo2018DeepAN}, which is caused by the difference between photographs and satellite images where the land cover types have random shapes like forest stands and bodies of water in contrast to the known shapes of ordinary objects. Kuo \textit{et al.} \cite{Kuo2018DeepAN} presents a method that is one of the leading results in the Deebglobe challenge, in which improving the performance relied on a variation of Deeplabv3+ \cite{Deeplabv3_plus} where ASPP (Atrous Spatial Pyramid Pooling) replace the fully connected layers of the ResNet backbone. Also, Deeplabv3+ uses an encoder-decoder architecture to reduce the effect of resolution loss due to pooling and strided convolution. The encoder-decoder method is a common approach in the leading methods of land cover segmentation because it preserves high-resolution features like texture and color, which play a significant role in distinguishing land covers. Hasan \textit{et al.} \cite{rs10060973} used state of the art semantic segmentation method, and added LIDAR data that improved upon the results a bit compared to other methods.

\subsubsection{Domain adaptation}
Domain adaptation reduces the domain shift between two different sets (source and target) with different distributions; it is achieved by aligning the distribution of one set to match the other, or by mapping both sets to a common space.

There are mainly three forms of domain adaptation approaches \cite{DBLP:journals/corr/abs-1802-03601} \cite{DBLP:journals/corr/abs-1812-02849}. The first method is a classical form of minimizing a distance between the source and the target data. Maximum Mean Discrepancy (MMD) is an example of minimization to achieve domain-invariant feature representation that performs well in both source and target domains. The second method is adversarial based domain adaptation that relies on the use of Generative Adversarial Networks (GANs) \cite{gan} to make one of the sets appear similar to the other one. Eric \textit{et al.} \cite{adda} is an example of the adversarial methods where the target data is translated to the source data with a discriminator to recognize between the two. The third method creates a shared representation for both domains not only to translate one domain to another but both domains to a common space, or have a transfer function capable of translating both models to each other and back to the original one. CycleGAN \cite{cycle} is an example of the third method where two discriminators are used to map images from the source data to the target data and vice-versa.

\subsubsection{Domain adaptation for semantic segmentation}
Domain adaptation is used for multiple applications, including semantic segmentation. All methods mentioned above were tested in the field of semantic segmentation with varying levels of success. In reality, domain adaptation has excellent potential for semantic segmentation. The reason stems from the lack of pixel annotation for images.

Generally, domain adaptation methods used for classification are not well translated for semantic segmentation \cite{Zhang2017CurriculumDA}. Therefore adversarial and reconstruction methods are preferred. Architectures such as FCNs in the Wild \cite{Hoffman2016FCNsIT}, and No More Discrimination \cite{No_More_Discrimination} are examples of adversarial domain adaptation that aims at using a GAN to generate source like images and then segment them using a network trained on the source data.
The reconstruction approach has been tested by many methods with different variations \cite{Hong_2018_CVPR} \cite{DBLP:journals/corr/abs-1903-12212} \cite{DBLP:journals/corr/abs-1802-10349} \cite{Vu_2019_CVPR}. The datasets that they used are almost exclusively the street-view images datasets, including Cityscape \cite{Cordts2016Cityscapes}, GTA5 \cite{Richter_2016_ECCV}, and Synthia \cite{Ros_2016_CVPR}. Chang \textit{et al.} \cite{Chang_2019_CVPR} is an example where a Domain Invariant Structure Extraction (DISE) framework was proposed to disentangle images into domain-invariant structure and domain-specific texture representations.
Li \textit{et al.} \cite{Li_2019_CVPR} proposed adding an extra step called bidirectional learning (BDL). The principle behind it is to add a bidirectional learning system and alternate between learning the segmentation and the image translation model using a loss that supervises the translation using the segmentation adaptation model. BDL prevents the translation model from converging to a point where the discriminator sees the images as being from the same distribution while not aligning the classes correctly, causing the segmentation to fail.

\section{Materials}

\subsection{Satellite data}\label{sat}
The satellite data used in the study is comprised of four EO satellites. The first one containing the most substantial bulk of the data is from the Sentinel-2 constellation of satellites. The second satellite is the WorldView-2 satellite containing fewer data. The third satellite is WorldView-3 through the DeepGlobe land cover dataset with images taken from WorldView-3 Vivid data \cite{Vivid}. Finally, Pleiades-1 is the fourth satellite data with the least amount of images.

Sentinel-2 constellation is composed of two polar-orbiting satellites placed in the same sun-synchronous orbit, phased at \ang{180} to each other. The satellites are named Sentinel-2A and Sentinel-2B. They each contain a multispectral sensor with 12 bands ranging from 60m/px resolution up to 10m/px, of which we used the RGB bands with 10m/px.

Through the European Space Agency's (ESA) Third Party Mission (TPM) data from WorldView-2 satellite is available for free as WorldView-2 European Cities \cite{esatpm}. WorldView-2 is a Very High Resolution (VHR) satellite that contains an 8-band multispectral sensor with 1.8m/px resolution shown in Table~\ref{worldview} and a 0.46cm/px pan-chromatic sensor. Similarly to the Sentinel-2's data, we only used the RGB bands.

Pleiades-1 is part of the Pleiades-1 constellation of satellites whose data is not available for the public for free. Nevertheless, we had access to a few rasters from which we extracted the RGB bands. It is similar in its properties to WorldView-2 (Table~\ref{pleiades1b}), which makes it interesting to see how a model trained on WorldView-2 would perform on the Pleiades-1 data.

The data obtained from satellite imagery comes in the form of raster images in a floating-point format. Both Sentinel-2 and WorldView-2 have a 12bit radiometric resolution encoded in a 32bit floating-point representation. These had to be encoded in an 8-bit unsigned integer format using QGIS, which is an open-source Geographical Information System (GIS) software. The main steps consisted of merging the rasters, translating the format to 8-bit images, and extracting the RGB channels before normalizing them to have a uniform illumination.
To form the datasets, we divided the obtained RGB rasters to patches of PNG images with $224\times224$ resolution for Sentinel-2,  $512\times512$ for WorldView-2, and $448\times448$ for Pleiades-1. The resulting number of images from Sentinel-2 is $37706$. The number of WorldView-2 images is $3570$. The Pleiades-1 dataset has the least number of images, which is $500$. The ratios for train, validation, and test sets for all the datasets are $80\%, 15\%,$, and $5\%$ respectively.

Additionally, to address the possibility of having clouds in the data, we acquired a set of Sentinel-2 and WorldView-2 rasters with cloud coverage. Sentinel-2 data come with level 1-C preprocessing, which includes cloud masks. WorldView-2 data, however, has no ground-truth cloud masks. 

For all the datasets, we introduced data augmentation during training as a mixture of random rotation and cropping.

\begin{table}[t!]
    \centering
    \caption{Sentinel2-B properties}
        \begin{tabular}{|m{1.4cm}|m{1.4cm}|m{1.4cm}|m{1.4cm}|}\hline
        Bands & Bandwidth ($nm$) & Central wavelength ($nm$) & Spacial resolution ($m$)\\\hline
        B2: Blue & 66 & 492.1 & 10 \\\hline
        B3: Green & 36 & 559 & 10 \\\hline
        B4: Red & 31 & 665 & 10 \\\hline
        \end{tabular}\\
    \label{sentinel2b}
    \end{table}

    \begin{table}[t!]
    \centering
    \caption{WorldView-2 properties}
        \begin{tabular}{|m{1.4cm}|m{1.4cm}|m{1.4cm}|m{1.4cm}|}\hline
        Bands & Bandwidth ($nm$) & Central wavelength ($nm$) & Spacial resolution ($m$)\\\hline
        B2: Blue & 60 & 480 & 1.84 \\\hline
        B3: Green & 70 & 545 & 1.84 \\\hline
        B5: Red & 60 & 660 & 1.84 \\\hline
        \end{tabular}\\
    \label{worldview}
    \end{table} 
    
    \begin{table}[t!]
    \centering
    \caption{Pleiades-1B properties}
        \begin{tabular}{|m{1.4cm}|m{1.4cm}|m{1.4cm}|m{1.4cm}|}\hline
        Bands & Bandwidth ($nm$) & Central wavelength ($nm$) & Spacial resolution ($m$)\\\hline
        B2: Blue & 120 & 490 & 2 \\\hline
        B3: Green & 120 & 550 & 2 \\\hline
        B4: Red & 120 & 660 & 2 \\\hline
        \end{tabular}\\
    \label{pleiades1b}
    \end{table} 

    \begin{table}[t!]
    \centering
    \caption{Classes in the label data}
        \begin{tabular}{|m{1.3cm}|P{1.3cm}|P{1.45cm}|}\hline
        Class name &Class code& Color \\\hline
        Unknown & 0 &\cellcolor{black}{} \\\hline
        Urban & 1 &\cellcolor{urban}{} \\\hline
        Agriculture & 2 &\contour{black}{\cellcolor{yellow}{}} \\\hline
        Rangeland & 3 &\cellcolor{rangeland}{} \\\hline
        Forestry & 4 &\cellcolor{green}{} \\\hline
        Water & 5 &\cellcolor{blue}{} \\\hline
        Barren & 6 &\contour{black}{\cellcolor{white}{}} \\\hline
        \end{tabular}\\
    \label{clc_table}
    \end{table}

\subsection{Labeled DeepGlobe data}
The DeepGlobe land cover segmentation dataset \cite{DBLP:journals/corr/abs-1805-06561} is used for comparison with available methods; it contains 1146 images with a resolution of $2000\times2000$, of which only 803 are labeled. The dataset is built from WorldView-3's vivid+ images \cite{Vivid}, and it is readily available without the need for preparation. It contains $12847$ images with a size of $612\times612$ pixels, which we cropped from the full size since using the full resolution images would require an excessive amount of GPU memory. The format of the images is 8-bit JPG with labels in PNG format. The images were divided into train, validation, and test subset with a ratio of $70\%, 20\%,$, and $10\%$ respectively. 

DeepGlobe dataset comes with its labels made by human annotators. The labels are the same ones shown in Table~\ref{clc_table} and were defined in \cite{DBLP:journals/corr/abs-1805-06561} as follows:
\begin{itemize}
    \item Urban land: Man-made, built up areas with human artifacts.
    \item Agriculture land: Farms, any planned (i.e. regular) plantation, cropland, orchards, vineyards, nurseries, and ornamental horticultural areas; confined feeding operations.
    \item Rangeland: Any non-forest, non-farm, green land, grass.
    \item Forest land: Any land with at least 20\% tree crown density plus clear cuts.
    \item Water: Rivers, oceans, lakes, wetland, ponds.
    \item Barren land: Mountain, rock, dessert, beach, land with no vegetation.
    \item Unknown: Clouds and others.
\end{itemize}

\subsection{Label data}\label{label}
The label data we used for the satellite imagery are obtained from Corine Land Cover by Copernicus (CLC) \cite{corinepdf}. CLC is a manually annotated pixel-based map of Europe with five major classes divided into a sum of forty-four sub-classes ranging from natural covers such as forests and water surfaces to human-made covers such as buildings and crops. Roughly, every six years, a new version of CLC has been available since the year 2000. The pixel resolution of CLC is 100m/px; however, there exists a version with 20m/px covering the whole area of Finland \cite{syke} and a 20m/px version covering Germany. We use these modified versions of CLC2018 for the satellite rasters captured between 2015 and 2017 and CLC2012 for the rasters captured from 2010 to 2012.

We applied some preprocessing on the labels by merging some classes to get them down to seven classes instead of the original forty-four to match the label data in the DeepGlobe dataset (see Table~\ref{clc_table}). The details of the classes are as follows:
\begin{itemize}
    \item Urban land: Man-made classes :Urban, Industrial, Mine.
    \item Agriculture land: Arable land, Perma crops, Pastures, Hereroagriculture.
    \item Rangeland: Shrubs, Inland wetland, Shrubs, Artificial non-agriculture green land
    \item Forest land: Forests. 
    \item Water: Inland water, Marine water.
    \item Barren land: Open spaces.
    \item Unknown: Clouds and others.
\end{itemize}
Corine's technical document \cite{corinepdf} provides a full description of each subclass. While the data covering Finland is accurate, the one covering Germany is not. Several errors can be spotted that ignore small objects such as individual houses in farms.

The label data were aligned to the same Coordinate Reference System (CRS) as the corresponding satellite rasters. Furthermore, we upsampled the CLC rasters to match the pixel resolution of the satellite rasters they cover; we then divided them into patches. Finally, we converted the label patches to single-channel 8-bit PNG images with each pixel's value ranging from 0 to 6 representing the corresponding class at that pixel.

\subsection{Study area}
\begin{figure}[t!]
            \begin{center}
                \includegraphics*[width=0.45\textwidth]{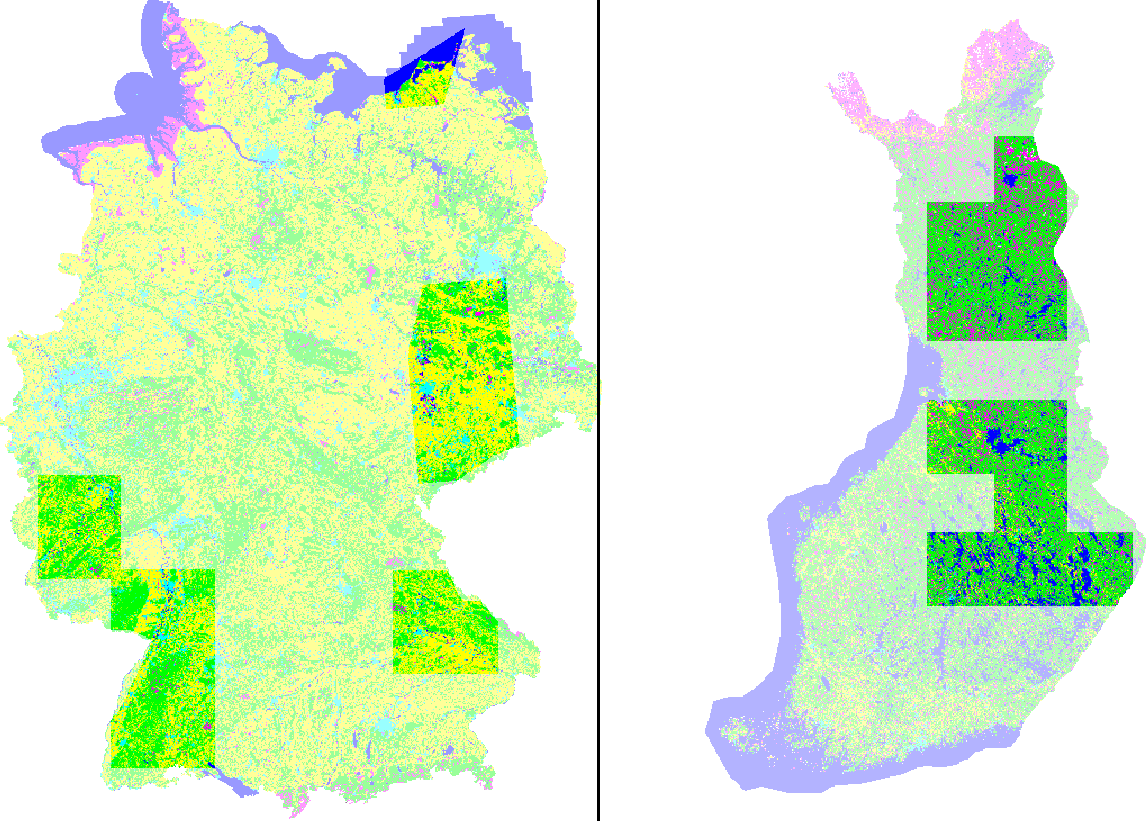}
            \end{center}
            \caption{Area covered by Sentinel-2. \textit{Left:} Area covered in Germany. \textit{Right:} Area covered in Finland}
            \label{fig:sentinel}
    \end{figure}

\begin{figure}[t!]
            \begin{center}
                \includegraphics*[width=0.45\textwidth]{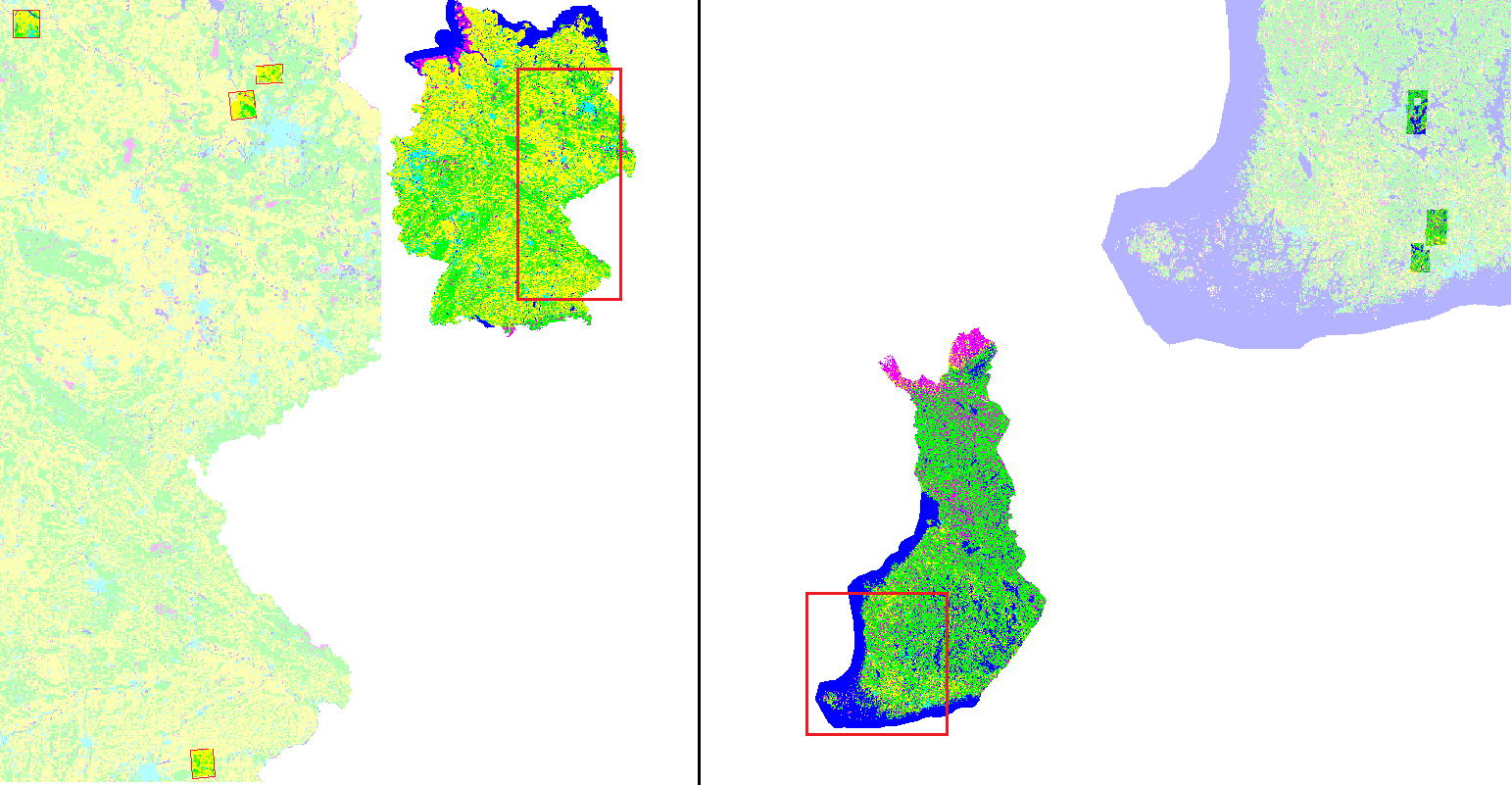}
            \end{center}
            \caption{Area covered by Worldview-2.\textit{Left:} Area covered in Germany. \textit{Right:} Area covered in Finland}
            \label{fig:WV}
    \end{figure}

The study area varies depending on the satellite. Sentinel-2 and WorldView-2  cover parts of Finland and Germany, with none of the areas between the satellites overlapping. WorldView-2 covers $1520.17km^2$ in Finland and $1310.18km^2$ in Germany (Fig~\ref{fig:WV}). The rasters were carefully chosen to avoid any cloud coverage that might compromise the efficiency of the training. Sentinel-2 covers a far larger area in Finland at around $128320.21km^2$ from all over the country. The area covered in Germany is also larger at around $74361.98km^2$ ( Fig~\ref{fig:sentinel} ). More data is used from Sentinel-2 because all of its data is available for free, whereas a limited amount of the WorldView-2 data is freely available. The Pleiades-1 data covers a small area of Finland at around $519.67km^2$ with no overlap data with the WorldView-2 data. Finally, the DeepGlobe dataset, which covers areas from India, Indonesia, and Thailand, corresponds to $1716.9km^2$.

It is characteristic of the data that Finland and Germany do not share much of the land cover distribution. The tree species for once are quite different, Finland has much more lakes and forests, Germany contains more urban areas and agriculture.

\section{Method}
The method we applied in this study uses domain adaptation to semantically segment unlabeled satellite images by land cover using a different labeled dataset. 

\subsection{Network architecture}
\begin{figure}[t!]
            \begin{center}
                \includegraphics*[width=0.48\textwidth]{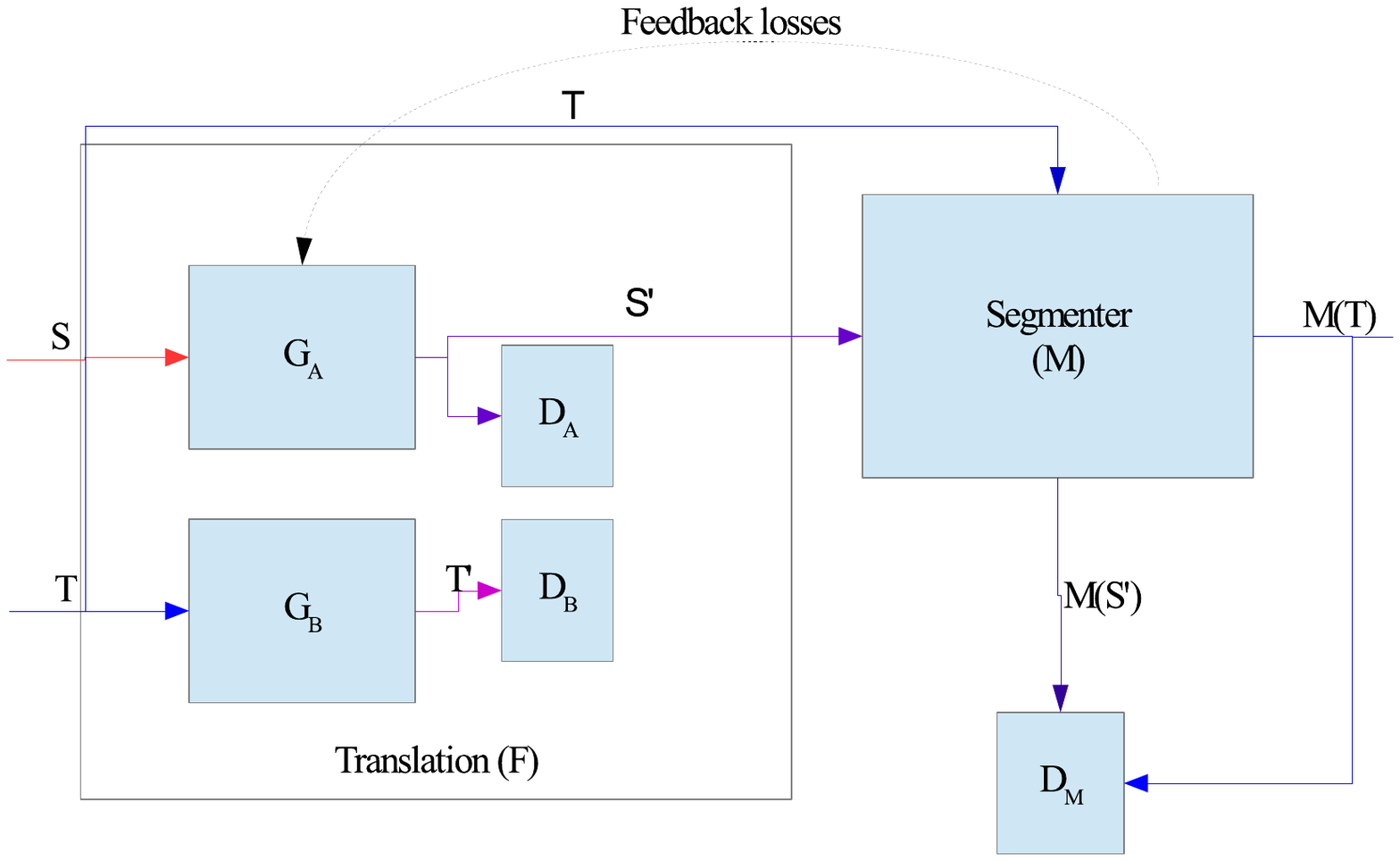}
            \end{center}
            \caption{Bidirectional learning architecture. S represents the source data, T is the Target data.}
            \label{fig:bdl}
    \end{figure}

The model used in this work is based on BDL\cite{Li_2019_CVPR} illustrated in Fig~\ref{fig:bdl}. It contains a translation network (F) that transforms source images to target images, a segmentation network (M) that assigns labels to the input images, and a domain discriminator (DM) that distinguishes between labels generated from target images and translated source images.

The architecture of the translation network (F) is a CycleGAN consisting of two 9-blocks ResNet generators and two 3-fully-connected-layers discriminators. The segmentation network is Deeplabv2, which is a network specialized for semantic segmentation, with ResNet101 as its backbone pre-trained with the ImageNet dataset. The domain discriminator is composed of 4 fully connected layers.

\subsection{Cloud masking network} \label{cloud}
\begin{figure}[t!]
            \begin{center}
                \includegraphics*[width=0.50\textwidth]{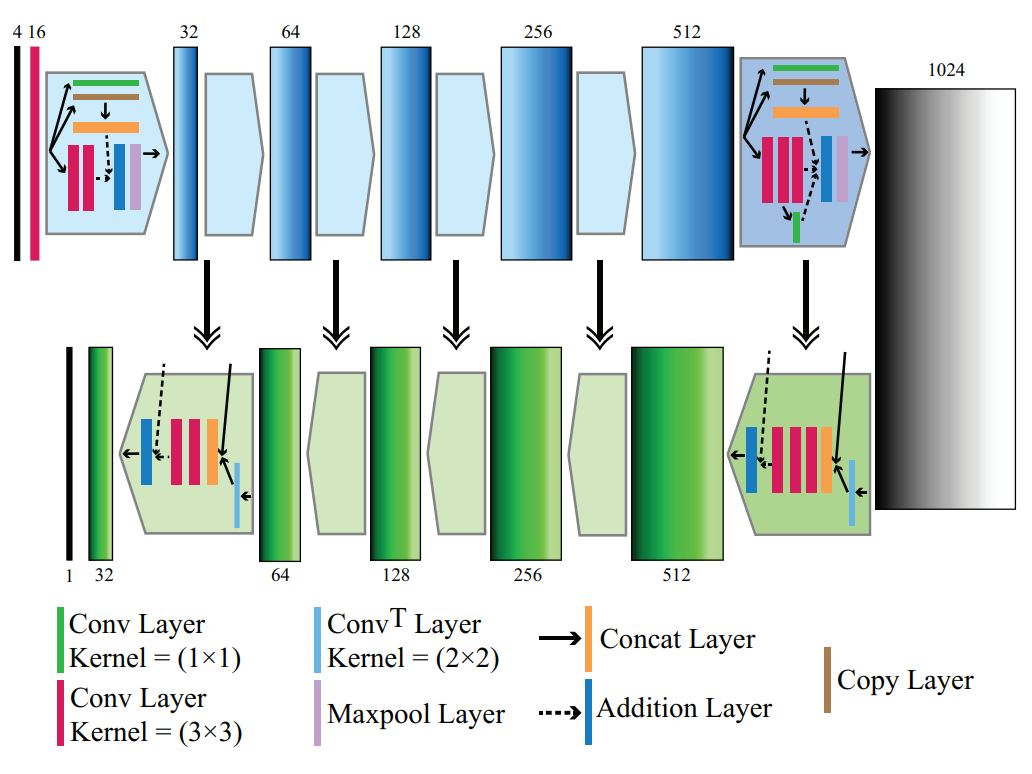}
            \end{center}
            \caption{Architecture of the cloud masking network used\cite{8898776}.}
            \label{fig:cld}
    \end{figure}

The cloud masking network we used for the cloud covered data is Cloud-Net \cite{8898776}. It consists of a U-net like architecture which contains six convolution blocks, and five deconvolution blocks (see Fig~\ref{fig:cld}). Cloud-Net is originally trained on four spectral layers from Landsat rasters (Red, Green, Blue, and Near Infra-Red).

\section{Experiments}

\subsection{Cloud masking}
\begin{figure}[t!]
            \begin{center}
                \includegraphics*[width=0.5\textwidth]{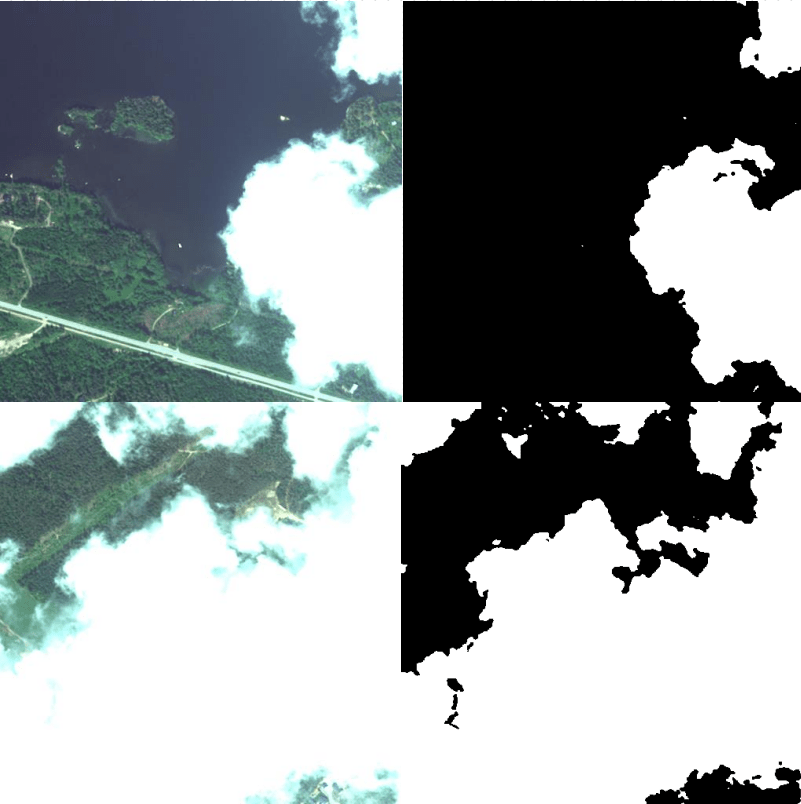}
            \end{center}
            \caption{Cloud masking of WorldView-2 images.}
            \label{fig:wvcld}
    \end{figure}

\begin{figure}[t!]
            \begin{center}
                \includegraphics*[width=0.47\textwidth]{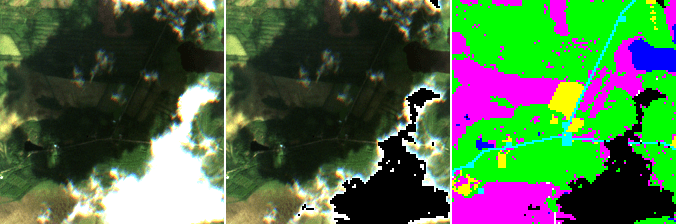}
            \end{center}
            \caption{Masked clouds from the GT Sentinel dataset. \textit{Left:} satellite image before cloud mas-2king. \textit{middle:} satellite image after cloud masking. \textit{right:} label image after cloud masking.}
            \label{fig:sensatcld}
    \end{figure}

\begin{figure}[t!]
            \begin{center}
                \includegraphics*[width=0.44\textwidth]{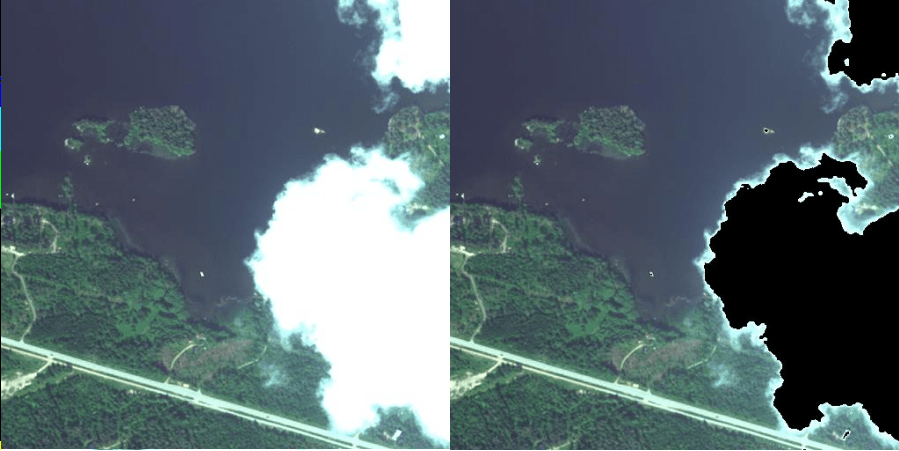}
            \end{center}
            \caption{Masked clouds from the Worldview-2 dataset. \textit{Left:} satellite image before cloud masking. \textit{right:} satellite image after cloud masking.}
            \label{fig:wvsatcld}
    \end{figure}

We first tried using the pretrained Cloud-Net \cite{8898776} model on the WorldView-2 data, but the results were not satisfactory. So, we trained the network using Sentinel-2's data, which has ground-truth cloud masks, and it yielded a good performance (Fig~\ref{fig:wvcld}).
To avoid adding a new class in the labels data, we merged the cloud masks with the unknown class (see Fig~\ref{fig:sensatcld}, Fig~\ref{fig:wvsatcld}).


\subsection{Training}
The training was performed on Nvidia GPUs (Tesla V100, Tesla P100, Tesla T4) \cite{v100} with 16GB of video memory for about $250,000$ iterations with a batch size of $4$. The batches were randomized for every iteration in the epoch. 
During the image translation phase, we resized the images to the lowest resolution between the source and target dataset. 

The training of the BDL network  uses the loss functions $l_M$ to train the segmentation network shown below:
\begin{equation}
l_M=\lambda_{adv}l_{adv}(M(S'),M(T))+l_{seg}(M(S'),Y_s) \label{M_loss}
\end{equation}
Where where $S$ is the source data and $T$ is the target data. $S'$ or $F(S)$ is the translated source to target data. $M(S')$, and $M(T)$ are the prediction labels for the translated source to target data and target data respectively. $Y_s$ is the GT label for the source data.  $l_{adv}$ is the adversarial loss of the domain discriminator:
\begin{equation}
l_{adv}(M(S'),M(T))=\mathbb{E}_{T}[D_M(M(T))]+\mathbb{E}_{S}[1-D_M(M(S))] \label{adv_loss}
\end{equation}
$\lambda_{adv}$ is the coefficient for the adversarial loss. $l_{seg}$ is the cross-entropy loss between GT labels and the predictions:
\begin{equation}
l_{seg}(M(S'),Y_s)=-\frac{1}{HW}\sum_{H,W} \sum_{c=1}^{C} 1_{[c={y_S}^{hw}]} log {P_S}^{hwc} \label{seg_loss}
\end{equation}
Where $C$ is the number of classes in the labels. $H$ and $W$ are the height and width of the label images respectively.

The corresponding loss for the translation network $l_F$ is: 
\begin{equation}
\begin{split}
l_F=\lambda_{GAN}(\mathbb{E}[\lambda_{D}D(T)+\mathbb{E}[1-\lambda_{D}D(S')]+\\
                  \mathbb{E}[\lambda_{D}D(T')+\mathbb{E}[1-\lambda_{D}D(S)])+\\
                    \lambda_{recon}[E[||F^{-1}(S')-S||_1]+E[||F^{-1}(T')-T||_1]]+\\
                    \lambda_{perA}\mathbb{E}[||M(S)-M(S')||_1]+\\ \lambda_{per\_recon}\mathbb{E}[||M(F^{-1}(S'))-M(S)||_1]+\\
                    \lambda_{perB}\mathbb{E}[||M(T)-M(T')||_1]+\\ \lambda_{per\_recon}\mathbb{E}[||M(F^{-1}(T'))-M(T)||_1] \label{F_loss}
\end{split}
\end{equation}
Where $T'$ or $F(T)$ is the translated target to source data. $M(S)$, and $M(T')$ are the prediction labels for the source data and the translated target to source data respectively. $F^{-1}$ is the inverse function of $F$.

$\lambda_{GAN}$ is the coefficient for the GAN loss. As for $\lambda_D$, it represents the coefficient for the discriminator loss. While $\lambda_{recon}$ is the coefficient for the reconstruction loss—or cyclic loss. $\lambda_{perA}$ signifies the coefficient scaling the perceptual loss of the source data, and $\lambda_{perB}$ the coefficient for the target data's perceptual loss. $\lambda_{per_recon}$ denotes the coefficient for the perceptual reconstruction loss. Those coefficients are the ones that will help guide the translation network using the segmentation network. 

\begin{figure}[t!]
            \begin{center}
                \includegraphics*[width=0.5\textwidth]{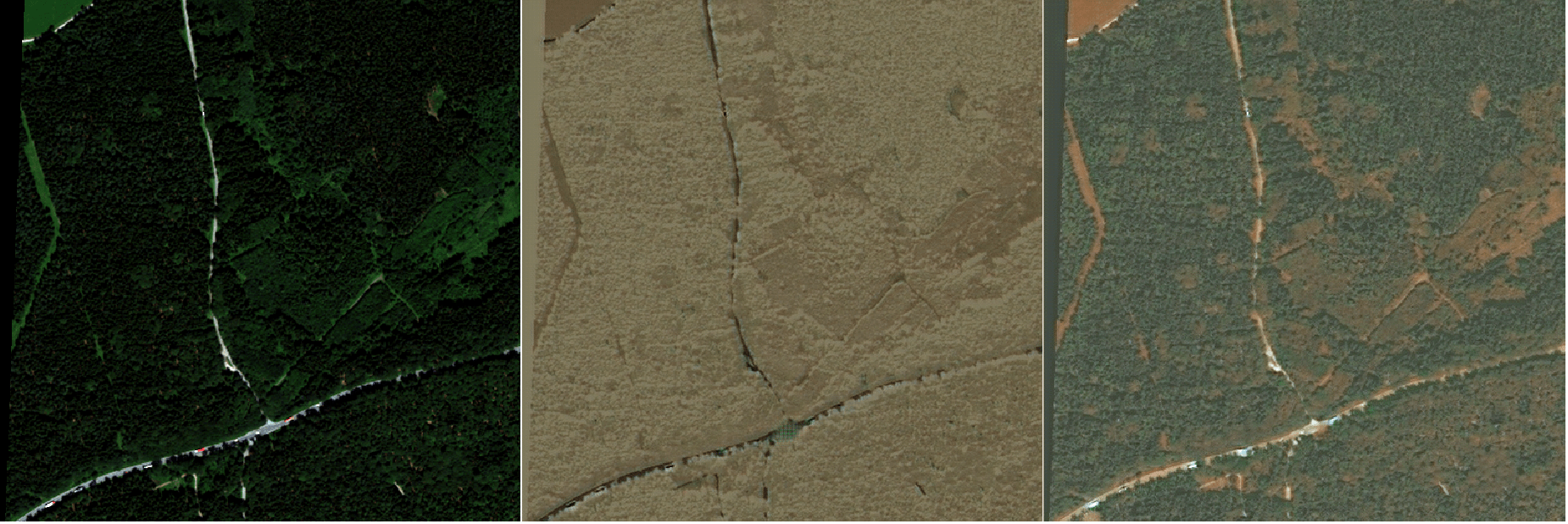}
            \end{center}
            \caption{Results of translation from WorldView-2 to DeepGlobe. \textit{Right:} result with $\lambda_{per}=0.1$. \textit{Center:} Worldview-2 image. \textit{Left:} WorldView-2 image.}
            \label{fig:noD_D}
    \end{figure}

\begin{figure}[t!]
            \begin{center}
                \includegraphics*[width=0.45\textwidth]{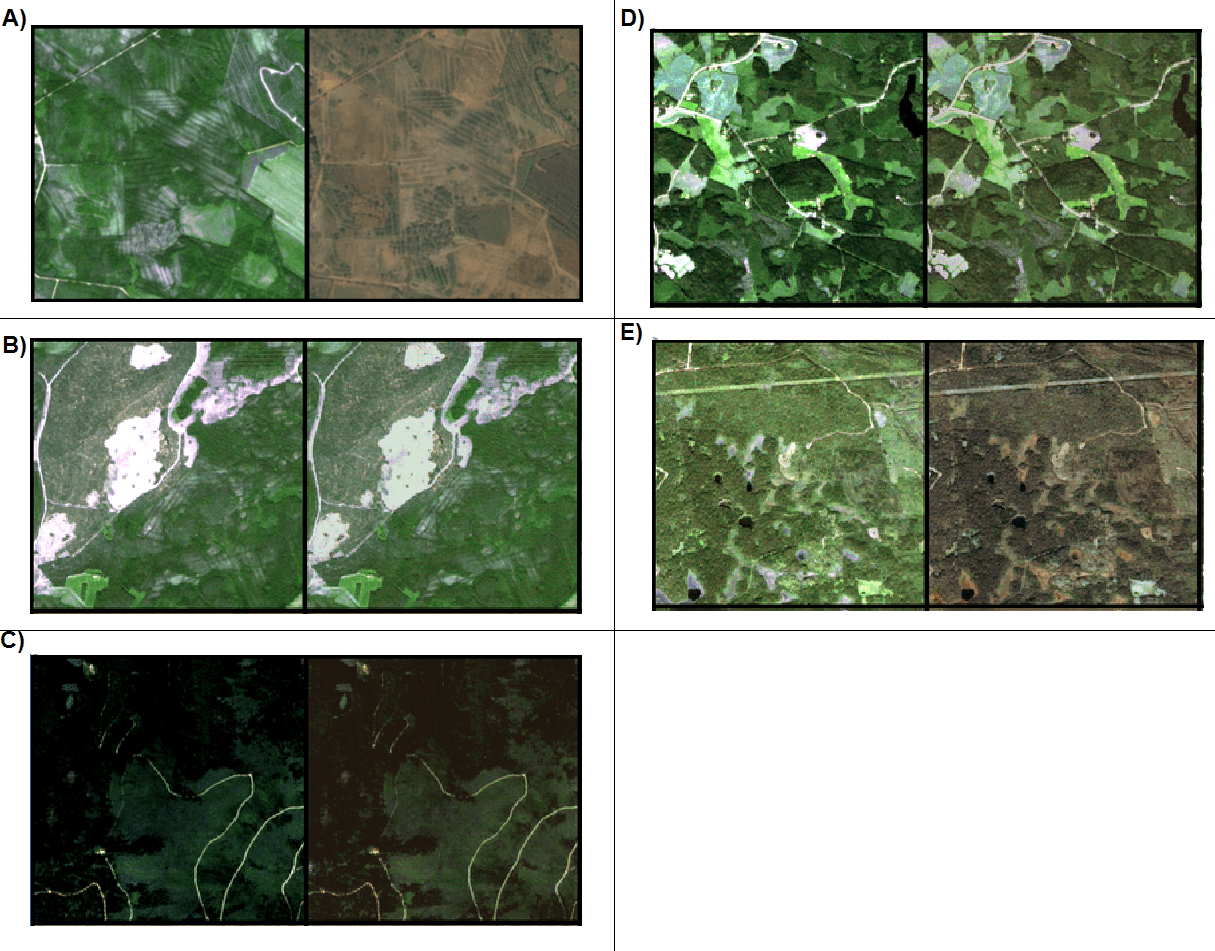}
            \end{center}
            \caption{Examples of multiple combinations of coefficients for Sentinel to DeepGlobe test In each example the left image is the Sentinel source and the left image is the translation to DeepGlobe. \textit{A:} $\lambda_{D}=1$, $\lambda_{perA}=0.1$, and $\lambda_{perB}=0.1$. \textit{A:} $\lambda_{D}=100$, $\lambda_{perA}=2$, and $\lambda_{perB}=0.5$. \textit{B:} $\lambda_{D}=1$, $\lambda_{perA}=10$, and $\lambda_{perB}=10$.
            \textit{C:} $\lambda_{D}=10$, $\lambda_{perA}=2$, and $\lambda_{perB}=0.5$. \textit{D:} $\lambda_{D}=50$, $\lambda_{perA}=2$, and $\lambda_{perB}=0.5$. \textit{E:} $\lambda_{D}=100$, $\lambda_{perA}=2$, and $\lambda_{perB}=0.5$.}
            \label{fig:sendg}
    \end{figure}

The coefficients presented above differentiate the original BDL network from what we used. Fig~\ref{fig:noD_D} shows an example where we compared $\lambda_{perA}=0.1$ with $\lambda_{perA}=1$, where the first case resulted in trees from WorldView-2 being replaced by the barren class in DeepGlobe domain while in the second case we obtained a more accurate translation. Each experiment required its own set of coefficients, so those are not fixed. We obtained those coefficients through trial and error. Another example is illustrated in Fig~\ref{fig:sendg}.


\subsection{Metrics}
In semantic segmentation literature, various metrics are used to measure the accuracy compared to the GT data. Those include Mean Intersection over Union (MIoU), Average Precision, Pixel Accuracy, and Boundary F1 score. In our experiments, we use MIoU which computes the mean of the rate of overlap between the GT segments and the resulting segmentation:
\begin{equation}
MIoU=\displaystyle{\frac{1}{n}}\sum_{i=1}^{n}\displaystyle{\frac{GT_i \cap Output_i}{GT_i \cup Output_i}} \label{miou}
\end{equation}
where $n$ is the number of classes. Another way to write the formula is:
\begin{equation} 
MIoU=\displaystyle{\frac{1}{n}}\sum_{i=1}^{n}\displaystyle{\frac{TP}{TP+FP+FN}} \label{miou1}
\end{equation}

\subsection{Baseline Results}
Since all datasets are labeled, it is possible to know the upper-bound results to compare them with those of domain adaptation. To obtain these results, we train and test the segmentation network on the target dataset. In this part, we used DeepGlobe, WorldView-2, WorldView-2 FI, and Pleiades-1 datasets. The results are shown in Table~\ref{upper}. 

    \begin{table*}[t!]
    \centering
    \caption{Upper-bound results}
        \begin{tabular}{|m{2.3cm}|m{1.2cm}|m{1.2cm}|m{1.2cm}|m{1.2cm}|m{1.2cm}|m{1.2cm}|m{1.2cm}|m{1.2cm}|}\hline
        Dataset & Unknown & Urban & Agriculture & Rangeland & Forest & Water & Barren & MIoU  \\\hline
        DeepGlobe & 0.0 & 0.02 & 57.55 & 0.0 & 0.01 & 0.56 & 0.0 & 50.42 \\\hline
        WorldView-2 & 0.0 & 43.99 & 69.81 & 4.99 & 39.76 & 55.79 & 0.0 & 57.96 \\\hline
        WorldView-2 FI & 0.0 & 48.47 & 60.67 & 2.39 & 36.79 & 30.42 & 0.0 & 50.89 \\\hline
        Pleiades-1 FI & 0.0 & 48.47 & 60.67 & 2.39 & 36.79 & 30.42 & 0.0 & 54.70 \\\hline
        \end{tabular}\\
    \label{upper}
    \end{table*}

\subsection{Results}

We refer to the test of using domain adaptation from WorldView-2 (Finland and Germany) to DeepGlobe as "\textit{WV2 to DG}".
Another test that adapts to DeepGlobe is Sentinel-2 to DeepGlobe, and we refer to it as "\textit{Sen to DG}".
To test how well domain adaptation performs between satellites when the location is similar, we implemented a test between Sentinel-2 and WorldView-2, and we refer to it as "\textit{Sen to WV2}".
The next experiment applies domain adaptation between two similar satellites, namely, as WorldView-2 and Pleiades-1 covering the same location, and we refer to it as "\textit{WV2FI to PLFI}". 
The final experiment aims at seeing how well domain adaptation works for different locations when the sensor used is the same. Therefore, we implemented WorldView-2 Germany to WorldView-2 Finland, and we refer to it as "\textit{WV2GR to WV2FI}".

To test the improvements over no domain adaptation, we performed a separate experiment with only the segmentation network enabled and ran the model on the target dataset's validation subset. As for the domain adaptation, we tested BDL using both Deeplabv2 and Deeplabv3+ as a backbone for the segmentation network.

\subsubsection{WV2 to DG}

    \begin{table*}[t!]
    \centering
    \caption{Results for WV2 to DG}
        \begin{tabular}{|m{2.3cm}|m{1.2cm}|m{1.2cm}|m{1.2cm}|m{1.2cm}|m{1.2cm}|m{1.2cm}|m{1.2cm}|m{1.2cm}|}\hline
        Network & Unknown & Urban & Agriculture & Rangeland & Forest & Water & Barren & MIoU  \\\hline
        No DA & 0.0 & 0.02 & 57.55 & 0.0 & 0.01 & 0.56 & 0.0 & 8.31 \\\hline
        DeeplabV2 DA & 0.0 & 43.99 & 63.81 & 4.99 & 39.76 & 55.79 & 0.0 & 29.76 \\\hline
        DeeplabV2 DA 2itr & 0.0 & 45.03 & 63.65 & 5.12 & 39.54 & 56.21 & 0.0 & 29.93 \\\hline
        DeeplabV3+ DA & 0.0 & 48.47 & 60.67 & 2.39 & 36.79 & 30.42 & 0.0 & 25.54 \\\hline
        \end{tabular}\\
    \label{wv2dg}
    \end{table*}

\begin{figure}[t!]
            \begin{center}
                \includegraphics*[width=0.41\textwidth]{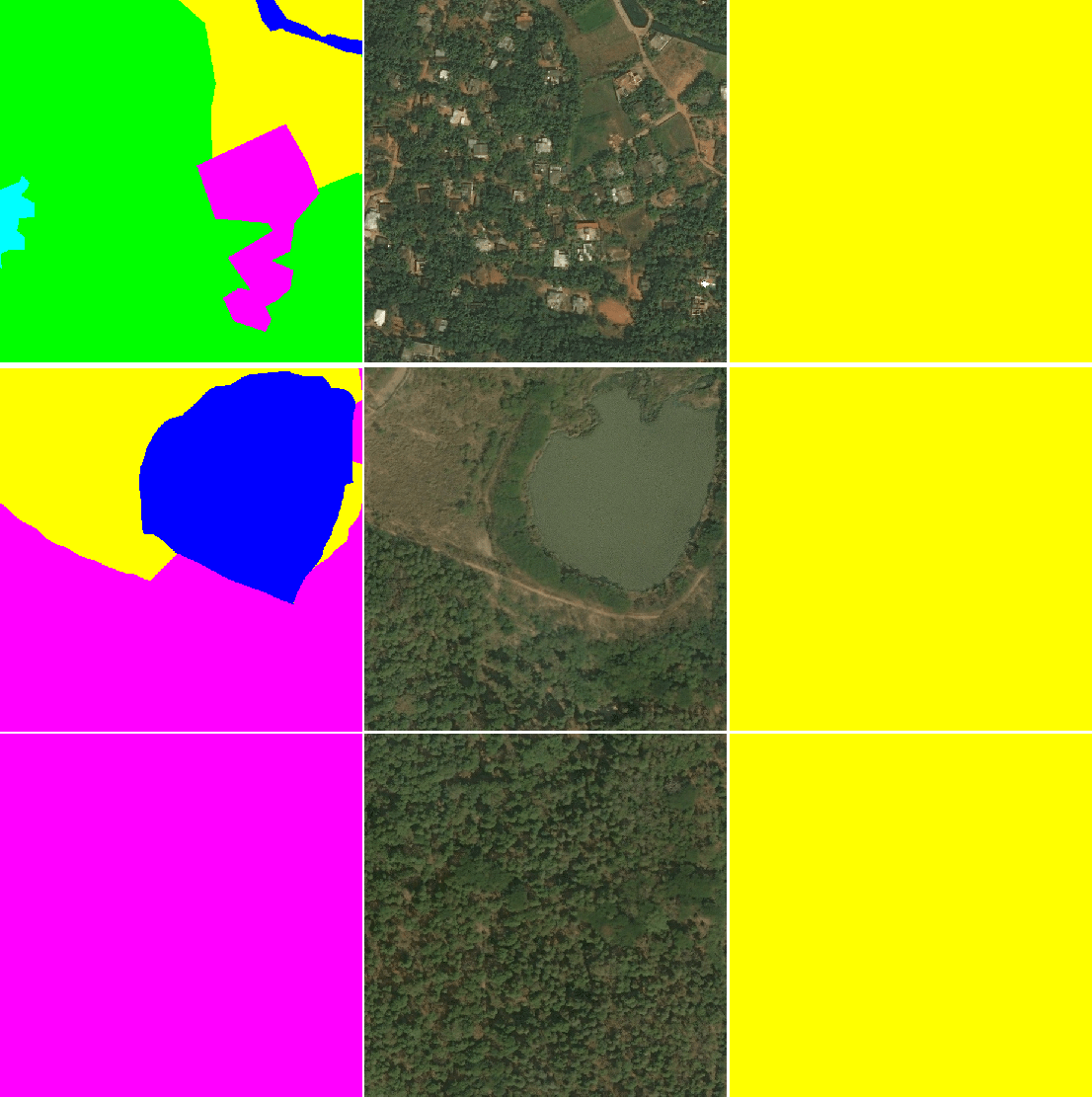}
            \end{center}
            \caption{Results of training without domain adaptation from WorldView-2 to DeepGlobe. \textit{Right:} output of model. \textit{middle:} test images from DeepGlobe dataset. \textit{Left:} ground truth images from DeepGlobe dataset.}
            \label{fig:WV2DGnocycle}
    \end{figure}

\begin{figure}[t!]
            \begin{center}
                \includegraphics*[width=0.41\textwidth]{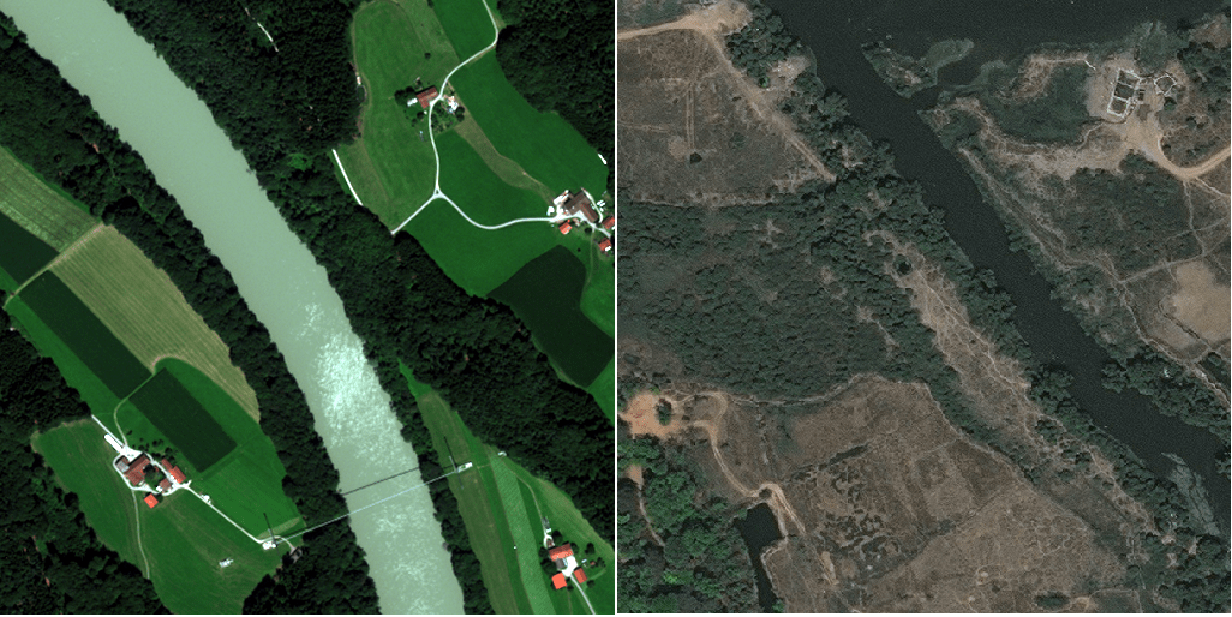}
            \end{center}
            \caption{Sample images from WorldView-2 dataset and DeepGlobe dataset. \textit{Right:} DeepGlobe dataset image. \textit{Left:} WorldView-2 dataset image.}
            \label{fig:wvdg}
    \end{figure}

The results of \textit{WV2 to DG} without domain adaptation are presented in Table~\ref{wv2dg}. The results are unsatisfactory, with many classes being different(Fig~\ref{fig:wvdg}) because images from WorldView-2 vary from images from DeepGlobe in both sensor properties and location. Fig~\ref{fig:WV2DGnocycle} shows an example of a few test images and the model output without domain adaptation. The network considers everything to be agriculture, which makes it very unreliable. 

\begin{figure}[t!]
            \begin{center}
                \includegraphics*[width=0.41\textwidth]{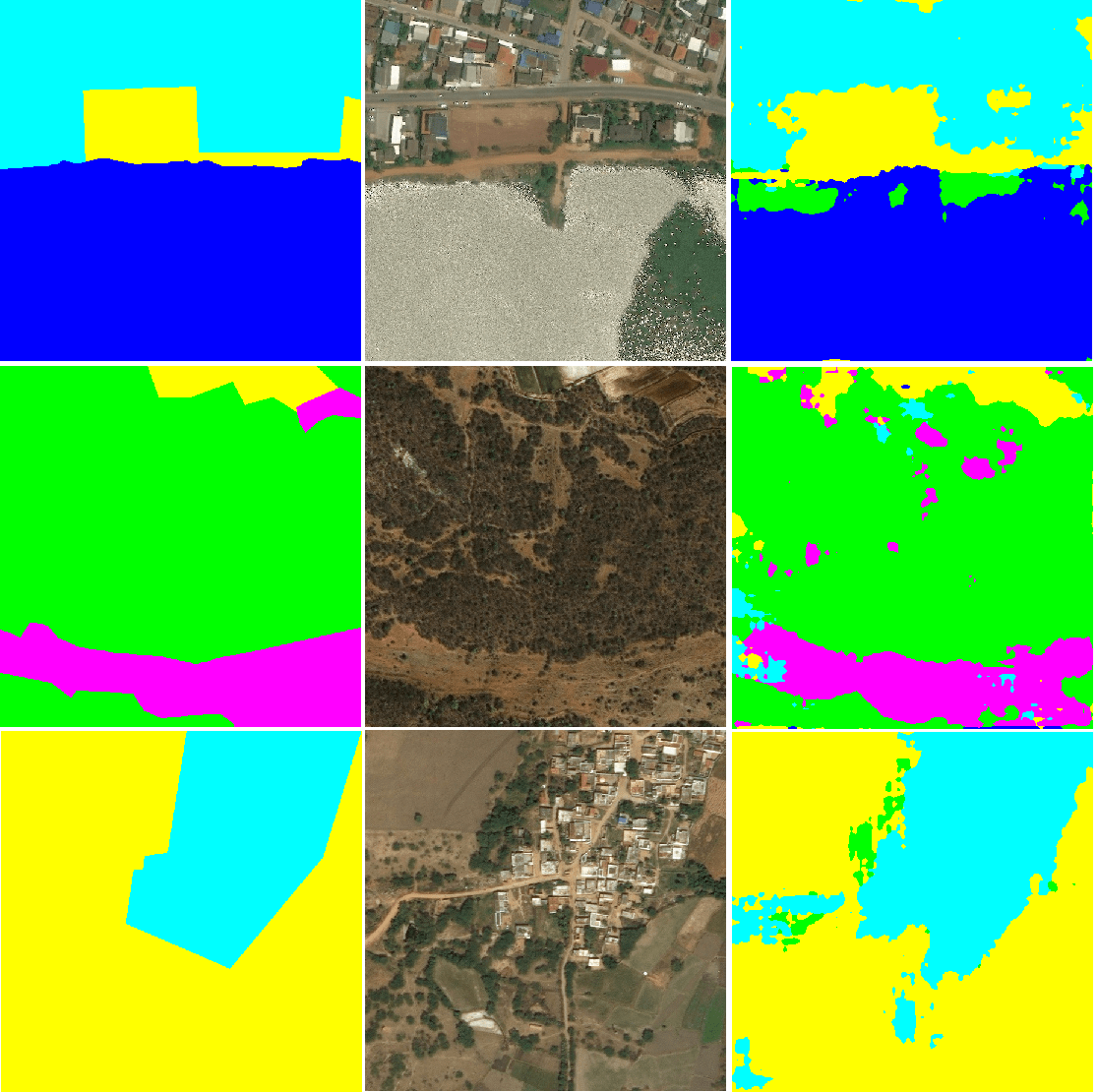}
            \end{center}
            \caption{Results of using domain adaptation from WorldView-2 to DeepGlobe. \textit{Right:} output of model. \textit{middle:} test images from DeepGlobe dataset. \textit{Left:} ground truth images from DeepGlobe dataset.}
            \label{fig:WV2DGcycle}
    \end{figure}

\begin{figure}[t!]
            \begin{center}
                \includegraphics*[width=0.41\textwidth]{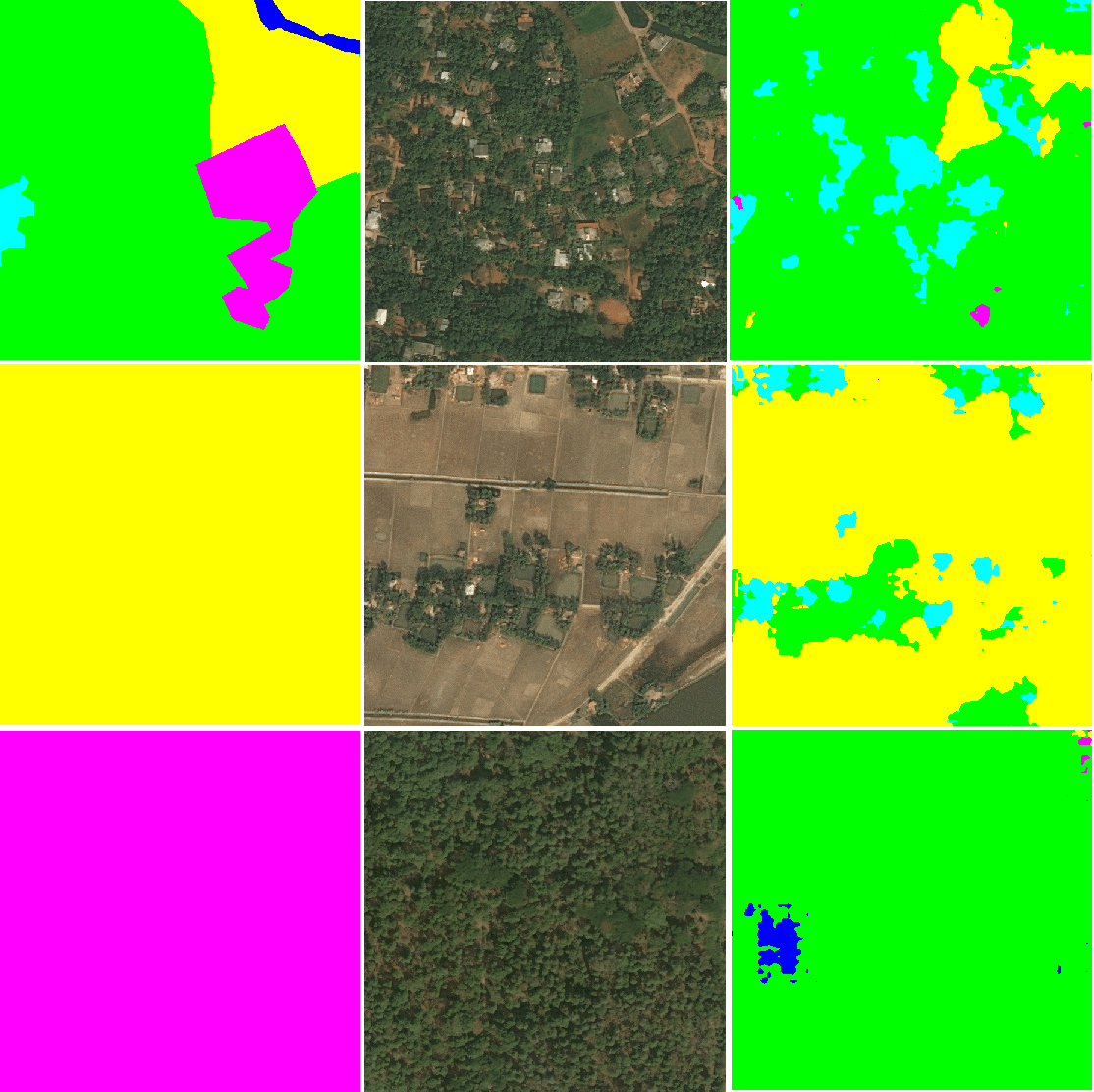}
            \end{center}
            \caption{Improving the ground truth using domain adaptation. \textit{Right:} output of model. \textit{middle:} test images from DeepGlobe dataset. \textit{Left:} ground truth images from DeepGlobe dataset.}
            \label{fig:WV2DGcycle2}
    \end{figure}

The results of using domain adaptation for \textit{WV2 to DG} are provided in Table~\ref{wv2dg}. Although the results are not very impressive numerically, there is a big difference between that and the MIoU results without domain adaptation going from 8.31 to 29.8. Fig~\ref{fig:WV2DGcycle} shows an example of model output after domain adaptation on a few test images, manifesting very similar labeling to the ground truth. Additionally, the results in some cases are better than the ground truth, which demonstrates that the GT annotation is imperfect. As an example, it is unclear what is considered forest and what is considered rangeland, and also, some small villages have been completely ignored in the ground truth while parts of them are not; this can observed in Fig~\ref{fig:WV2DGcycle2}. 


\subsubsection{Sen to DG}  

    \begin{table*}[t!]
    \centering
    \caption{Results for Sen to DG}
        \begin{tabular}{|m{2.3cm}|m{1.2cm}|m{1.2cm}|m{1.2cm}|m{1.2cm}|m{1.2cm}|m{1.2cm}|m{1.2cm}|m{1.2cm}|}\hline
        Network & Unknown & Urban & Agriculture & Rangeland & Forest & Water & Barren & MIoU  \\\hline
        No DA & 0.0 & 9.29 & 52.19 & 8.02 & 19.66 & 28.74 & 0.11 & 16.86 \\\hline
        DeeplabV2 DA & 0.1 & 29.78 & 40.42 & 9.73 & 23.3 & 62.67 & 0.58 & 23.8 \\\hline
        DeeplabV2 DA 2itr & 0.0 & 31.27 & 43.18 & 10.66 & 24.11 & 65.24 & 0.36 & 24.92 \\\hline
        DeeplabV3+ DA & 0.0 & 34.76 & 32.93 & 7.93 & 28.55 & 49.61 & 0.92 & 22.1 \\\hline
        \end{tabular}\\
    \label{sen2dg}
    \end{table*}
Unlike \textit{WV2 to DG}, where WorldView-2 is a satellite of somewhat similar properties with the main difference being the location, \textit{Sen to DG} is trying to perform domain adaptation between two very different satellites.

\begin{figure}[t!]
            \begin{center}
                \includegraphics*[width=0.41\textwidth]{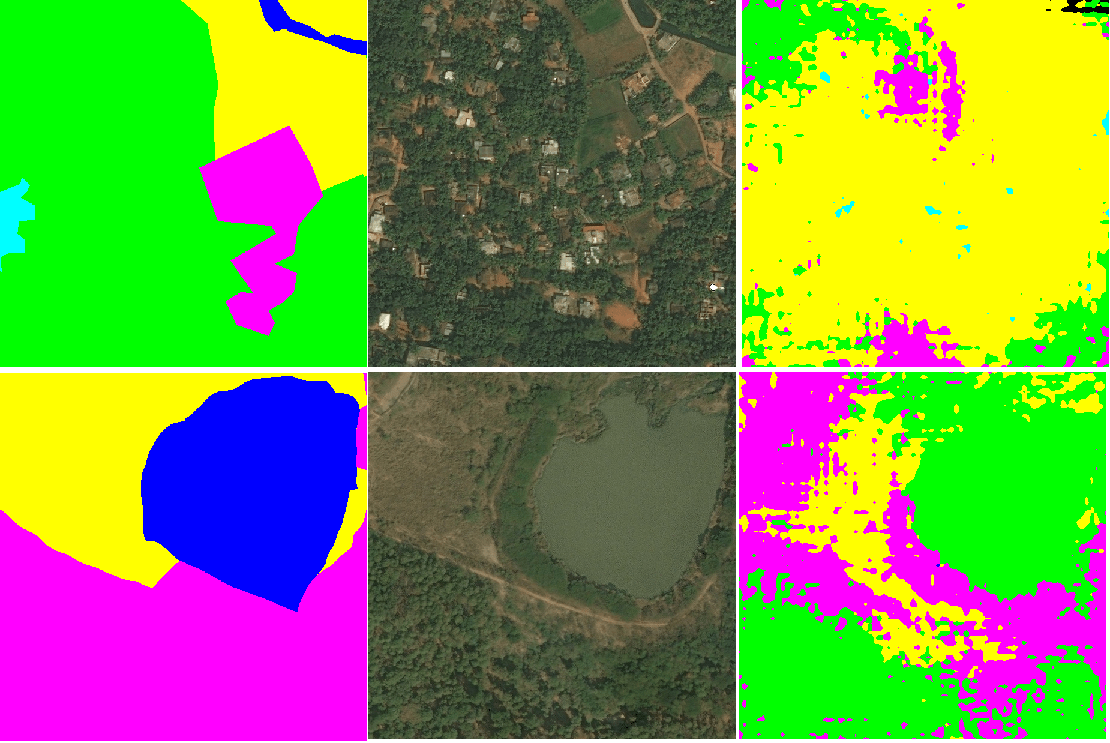}
            \end{center}
            \caption{Results of training without domain adaptation from Sentinel-2 to DeepGlobe. \textit{Right:} output of model. \textit{middle:} test images from DeepGlobe dataset. \textit{Left:} ground truth images from DeepGlobe dataset.}
            \label{fig:Sen2DGnocycle}
    \end{figure}

The results without domain adaptation are not reliable, albeit better than with \textit{WV2 to DG} since Sentinel-2 contains considerably more data. However, it is not enough to produce acceptable results, as illustrated in Fig~\ref{fig:Sen2DGnocycle}.  

\begin{figure}[t!]
            \begin{center}
                \includegraphics*[width=0.41\textwidth]{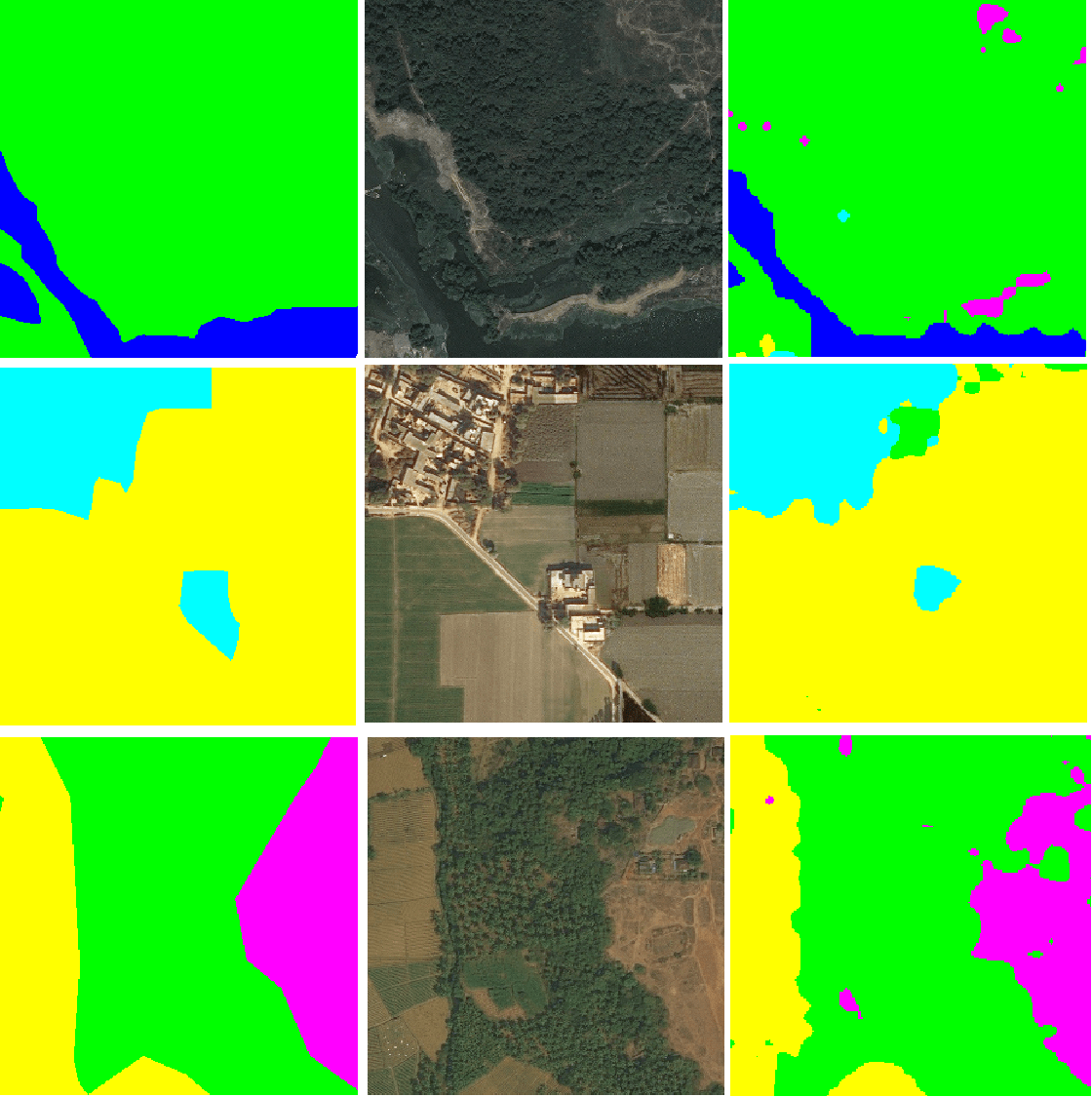}
            \end{center}
            \caption{Results of training with domain adaptation from Sentinel-2 to DeepGlobe. \textit{Right:} output of model. \textit{middle:} test images from DeepGlobe dataset. \textit{Left:} ground truth images from DeepGlobe dataset.}
            \label{fig:sen2dgcycle}
    \end{figure}

Using domain adaptation improved the results from a MIoU of 16.86 to 23.8, as shown in Table~\ref{sen2dg}. When compared to the state-of-the-art results obtained with Deeplabv2 on DeepGlobe, that is 52.24 \cite{Tian_2018_CVPR_Workshops}, our results are modest, but visually they are still good as illustrated in Fig~\ref{fig:sen2dgcycle} with a few examples. The results are not as good as with \textit{WV2 to DG}, which could be explained with the considerable difference of pixel resolution between Sentinel-2 and WorldView-3. However, this is still a good step forward as the Sentinel-2 data is free, whereas WorldView-3 data is not.


\subsubsection{Sen to WV2}  

    \begin{table*}[t!]
    \centering
    \caption{Results for Sen to WV2}
        \begin{tabular}{|m{2.3cm}|m{1.2cm}|m{1.2cm}|m{1.2cm}|m{1.2cm}|m{1.2cm}|m{1.2cm}|m{1.2cm}|m{1.2cm}|}\hline
        Network & Unknown & Urban & Agriculture & Rangeland & Forest & Water & Barren & MIoU  \\\hline
        No DA & 14.24 & 11.33 & 41.58 & 2.92 & 50.28 & 58.21 & 1.2 & 25.7 \\\hline
        DeeplabV2 DA & 67.71 & 34.65 & 79.87 & 6.16 & 76.27 & 77.06 & 0.0 & 48.82 \\\hline
        DeeplabV2 DA 2itr & 69.53 & 31.5 & 81.32 & 7.31 & 76.11 & 80.79 & 0.0 & 49.49 \\\hline
        DeeplabV3+ DA & 68.08 & 36.94 & 80.27 & 9.79 & 75.78 & 76.38 & 0.0 & 49.61 \\\hline
        \end{tabular}\\
    \label{sen2wv}
    \end{table*}
    
\begin{figure}[t!]
            \begin{center}
                \includegraphics*[width=0.41\textwidth]{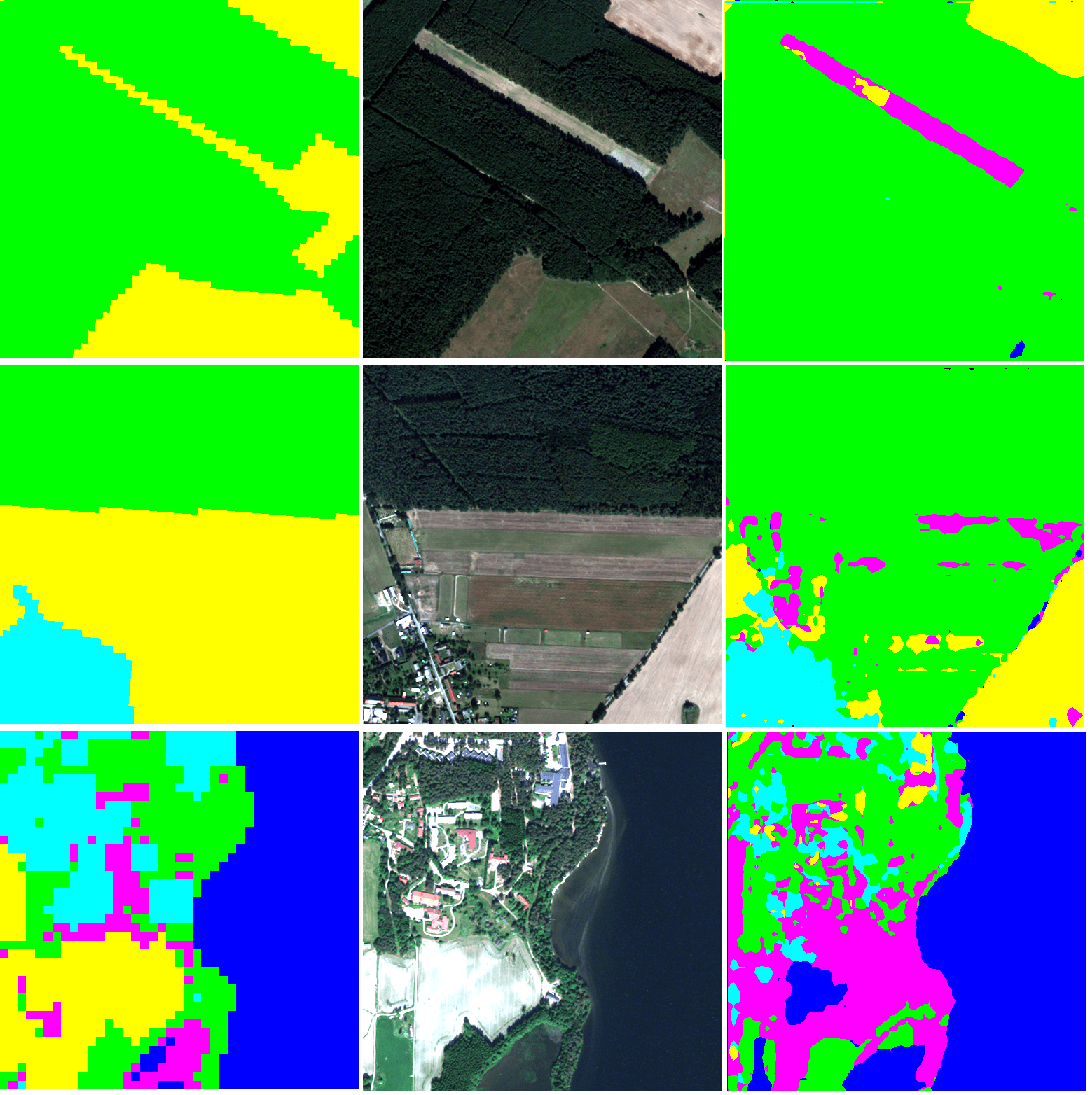}
            \end{center}
            \caption{Results of training without domain adaptation from Sentinel-2 to WorldView-2. \textit{Right:} output of model. \textit{middle:} test images from WorldView-2 dataset. \textit{Left:} ground truth images from WorldView-2 dataset.}
            \label{fig:Sen2WVnocycle}
    \end{figure}

\begin{figure}[t!]
            \begin{center}
                \includegraphics*[width=0.41\textwidth]{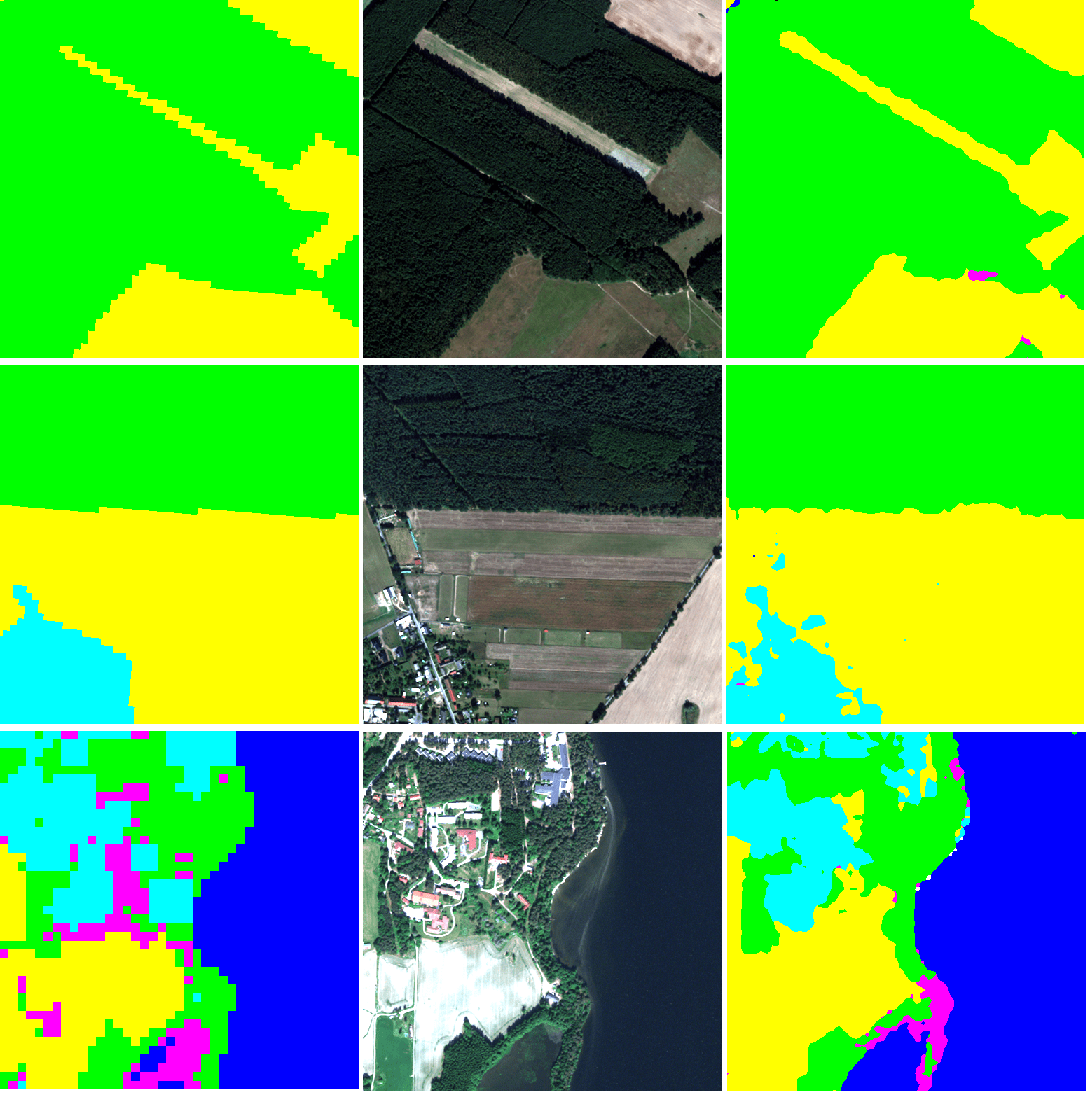}
            \end{center}
            \caption{Results of training with domain adaptation from Sentinel-2 to WorldView-2. \textit{Right:} output of model. \textit{middle:} test images from WorldView-2 dataset. \textit{Left:} ground truth images from WorldView-2 dataset.}
            \label{fig:Sen2WVcycle}
    \end{figure}

\begin{figure}[t!]
            \begin{center}
                \includegraphics*[width=0.41\textwidth]{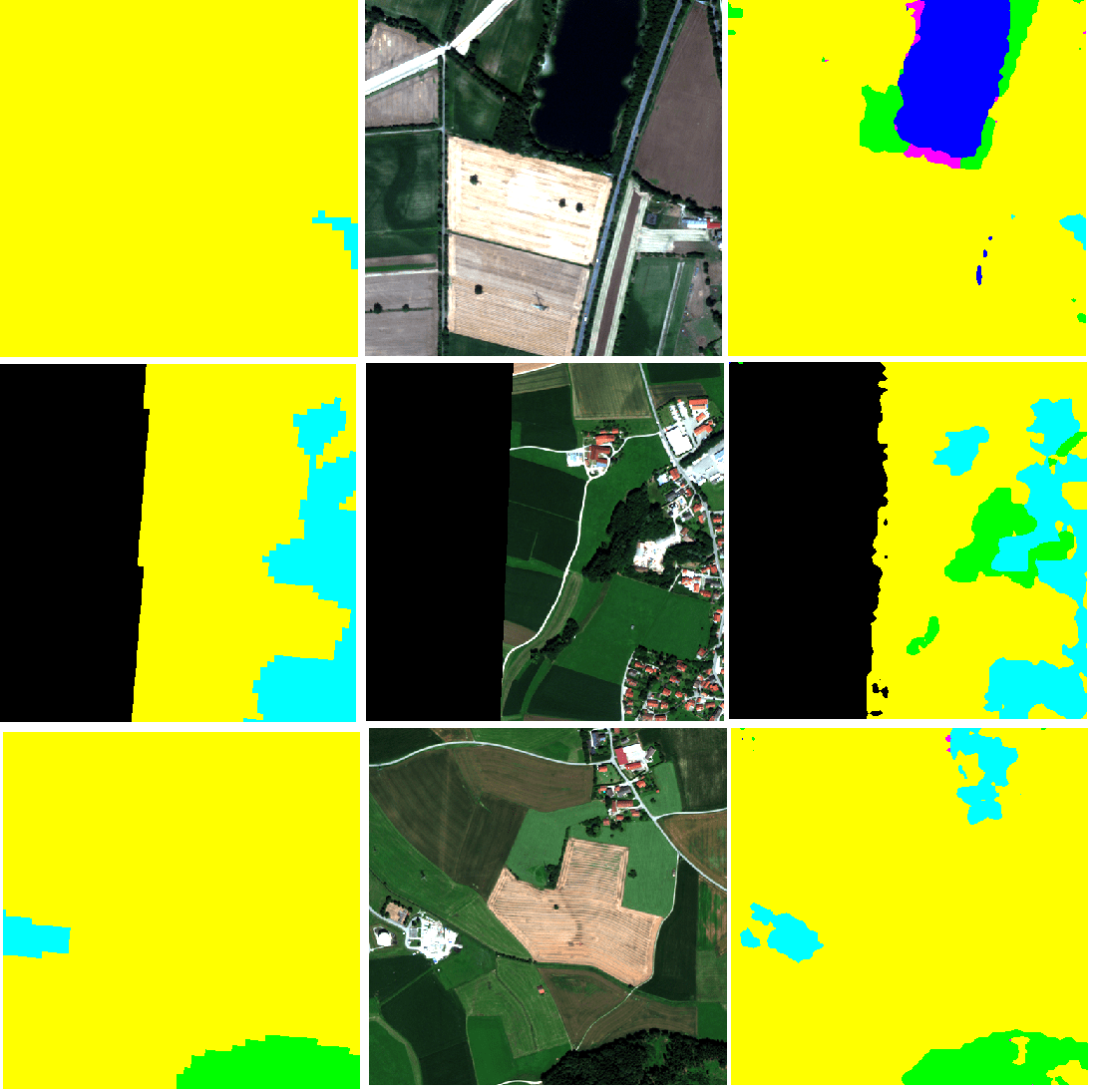}
            \end{center}
            \caption{Results of training with domain adaptation from Sentinel-2 to WorldView-2. \textit{Right:} output of model. \textit{middle:} test images from WorldView-2 dataset. \textit{Left:} ground truth images from WorldView-2 dataset.}
            \label{fig:Sen2WVcycle2}
    \end{figure}

The \textit{Sen to WV2} results without domain adaptation have some inaccuracy, especially mixing up similarly looking classes such as forestry and agriculture when seen in different resolutions. Examples can be found in Fig~\ref{fig:Sen2WVnocycle}, highlighting the errors mentioned. 
Using domain adaptation resulted in a significant MIoU improvement compared to the ones obtained without it, as presented in Table~\ref{sen2wv}. Samples from the results can be seen in Fig~\ref{fig:Sen2WVcycle}. As mentioned in Section~\ref{label}, the ground truth labels for Germany lack precision, which would limit the performance even when training on that specific dataset. However, having more accurate labels covering in Finland helps correct the errors in the German ones ending up with better results than the GT in a few cases, as displayed in Fig~\ref{fig:Sen2WVcycle2}.

\begin{figure}[t!]
            \begin{center}
                \includegraphics*[width=0.41\textwidth]{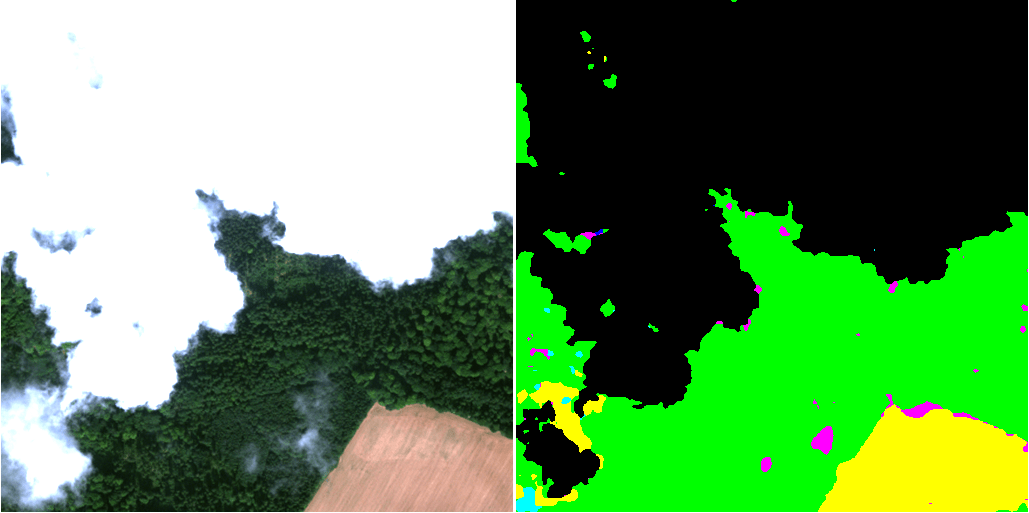}
            \end{center}
            \caption{Results of training with domain adaptation from Sentinel-2 with clouds masked to WorldView-2 with clouds masked. \textit{Right:} output of model. \textit{left:} test images from WorldView-2 dataset.}
            \label{fig:Sen2WVccycle}
    \end{figure}

In the cloud masking test, there are no ground-truth labels for the WorldView-2 images, so it is not possible to have a correct MIoU value. However, by overlaying the cloud masks generated by the network in Section~\ref{cloud} on the ground truth CLC labels, we get $43.7\%$ MIoU. An example of the results is shown in Fig~\ref{fig:Sen2WVccycle}.


\subsubsection{WV2FI to PLFI}   

    \begin{table*}[t!]
    \centering
    \caption{Results for WV2FI to PLFI}
        \begin{tabular}{|m{2.3cm}|m{1.2cm}|m{1.2cm}|m{1.2cm}|m{1.2cm}|m{1.2cm}|m{1.2cm}|m{1.2cm}|m{1.2cm}|}\hline
        Network & Unknown & Urban & Agriculture & Rangeland & Forest & Water & Barren & MIoU  \\\hline
        No DA & 8.90 & 40.15 & 52.73 & 2.74 & 59.85 & 9.27 & 0.0 & 24.8 \\\hline
        DeeplabV2 DA & 0.0 & 49.64 & 80.85 & 21.14 & 76.71 & 6.81 & 0.04 & 33.6 \\\hline
        DeeplabV2 DA 2itr & 0.0 & 51.45 & 80.40 & 22.82 & 75.5 & 6.98 & 0.8 & 34.0 \\\hline        
        DeeplabV3+ DA & 0.0 & 49.9 & 79.82 & 21.61 & 76.39 & 5.76 & 0.7 & 33.36 \\\hline
        \end{tabular}\\
    \label{wv2pl}
    \end{table*}
    
\begin{figure}[t!]
            \begin{center}
                \includegraphics*[width=0.41\textwidth]{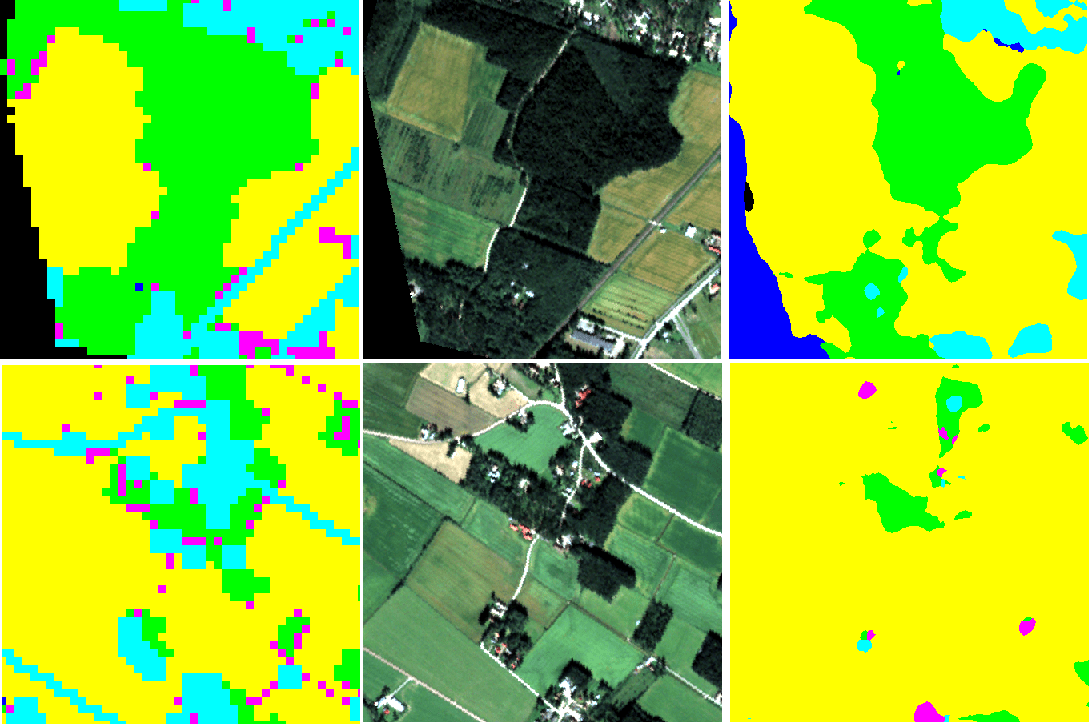}
            \end{center}
            \caption{Results of training without domain adaptation from WorldView-2 Finland to Pleiades-1. \textit{Right:} output of model. \textit{middle:} test images from Pleiades-1 dataset. \textit{Left:} ground truth images from Pleiades-1 dataset.}
            \label{fig:WVFI2PLFInocycle}
    \end{figure}

The results of \textit{WV2FI to PLFI} without using domain adaptation is not as good as expected. Even though the differences in sensors of both satellites are not significant, the results do not translate well. Fig~\ref{fig:WVFI2PLFInocycle} shows an example of the results of \textit{WV2FI to PLFI} without domain adaptation.

\begin{figure}[t!]
            \begin{center}
                \includegraphics*[width=0.41\textwidth]{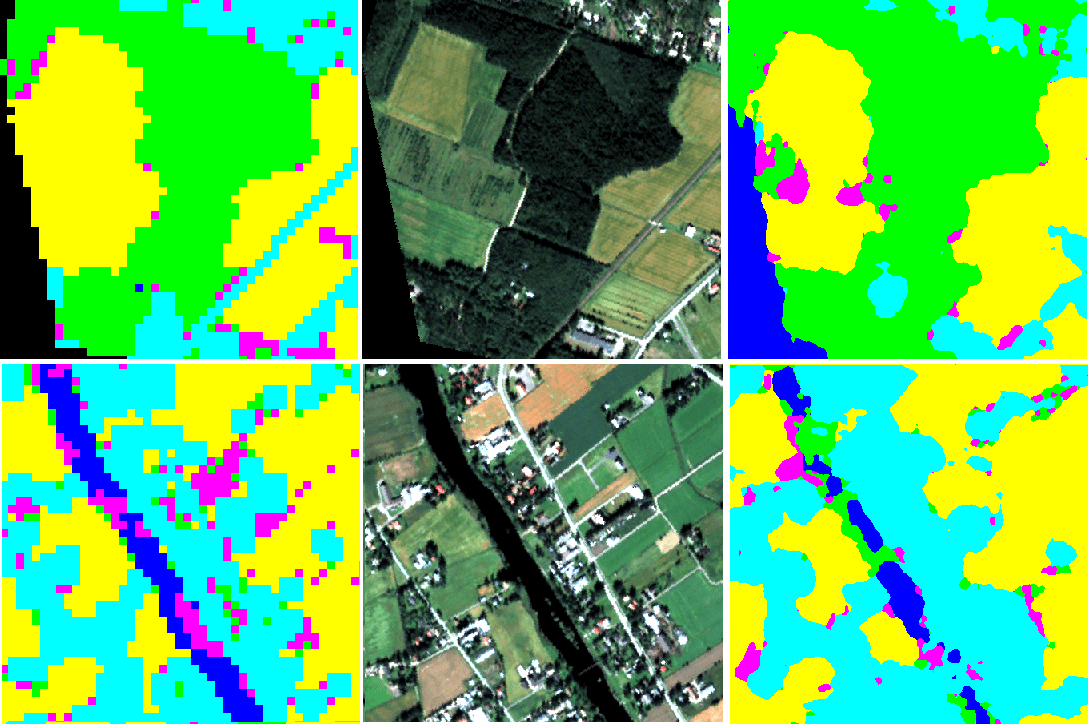}
            \end{center}
            \caption{Results of training with domain adaptation from WorldView-2 (Finland) to Pleiades-1 (Finland). \textit{Right:} output of model. \textit{middle:} test images from Pleiades-1 (Finland) dataset. \textit{Left:} ground truth images from Pleiades-1 (Finland) dataset.}
            \label{fig:WVFI2PLFIcyclev3}
    \end{figure}

With a limited amount of data like the case with the Pleiades-1 dataset, training a DNN on this data would lead to overfitting. Therefore, applying domain adaptation is an excellent way to test the efficiency of such a method to perform land mapping on a small dataset. \textit{WV2FI to PLFI} shows good results with almost a 10\% increase in MIoU, as illustrated in Fig~\ref{fig:WVFI2PLFIcyclev3} which is promising when considering that it would be costly to train on Pleiades-1 data. The detailed results are presented in Table~\ref{wv2pl}.


\subsubsection{WV2GR to WV2FI}   

    \begin{table*}[t!]
    \centering
    \caption{Results for WV2GR to WV2FI}
        \begin{tabular}{|m{2.3cm}|m{1.2cm}|m{1.2cm}|m{1.2cm}|m{1.2cm}|m{1.2cm}|m{1.2cm}|m{1.2cm}|m{1.2cm}|}\hline
        Network & Unknown & Urban & Agriculture & Rangeland & Forest & Water & Barren & MIoU  \\\hline
        No DA & 0.0 & 4.47 & 17.63 & 0.05 & 5.07 & 37.51 & 0.0 & 9.25 \\\hline
        DeeplabV2 DA & 0.0 & 23.9 & 39.91 & 2.68 & 71.07 & 51.84 & 0.0 & 27.06 \\\hline
        DeeplabV2 DA 2itr & 0.0 & 23.79 & 40.02 & 2.55 & 70.64 & 51.88 & 0.0 & 26.98 \\\hline
        DeeplabV3+ DA & 0.0 & 24.92 & 39.62 & 3.94 & 70.56 & 57.63 & 0.0 & 28.09 \\\hline
        \end{tabular}\\
    \label{wv2wv}
    \end{table*} 
    
Finally, the results of \textit{WV2GR to WV2FI} without domain adaptation are surprisingly weak when considering that the satellite is the same, and the labels were both based on CLC. Table~\ref{wv2wv} shows that the MIoU is only 9.25 MIoU, which is the second-lowest score in the list of tests ran. A sample of those results is shown in Fig~\ref{fig:WVFI2PLFInocycle}.

\begin{figure}[t!]
            \begin{center}
                \includegraphics*[width=0.41\textwidth]{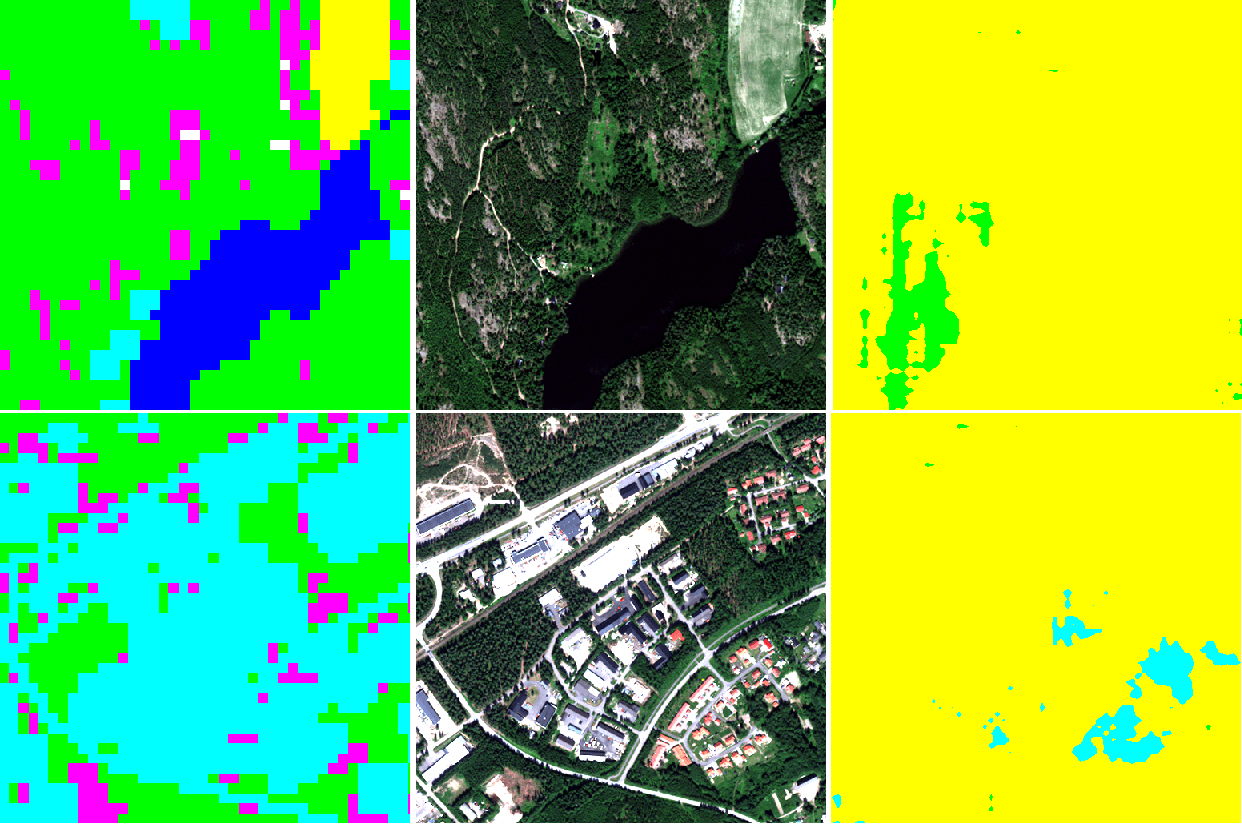}
            \end{center}
            \caption{Results of training without domain adaptation from WorldView-2 (Germany) to WorldView-2 (Finland). \textit{Right:} output of model. \textit{middle:} test images from WorldView-2 (Finland) dataset. \textit{Left:} ground truth images from WorldView-2 (Finland) dataset.}
            \label{fig:WVGR2WVFInocycle}
    \end{figure}

\begin{figure}[t!]
            \begin{center}
                \includegraphics*[width=0.41\textwidth]{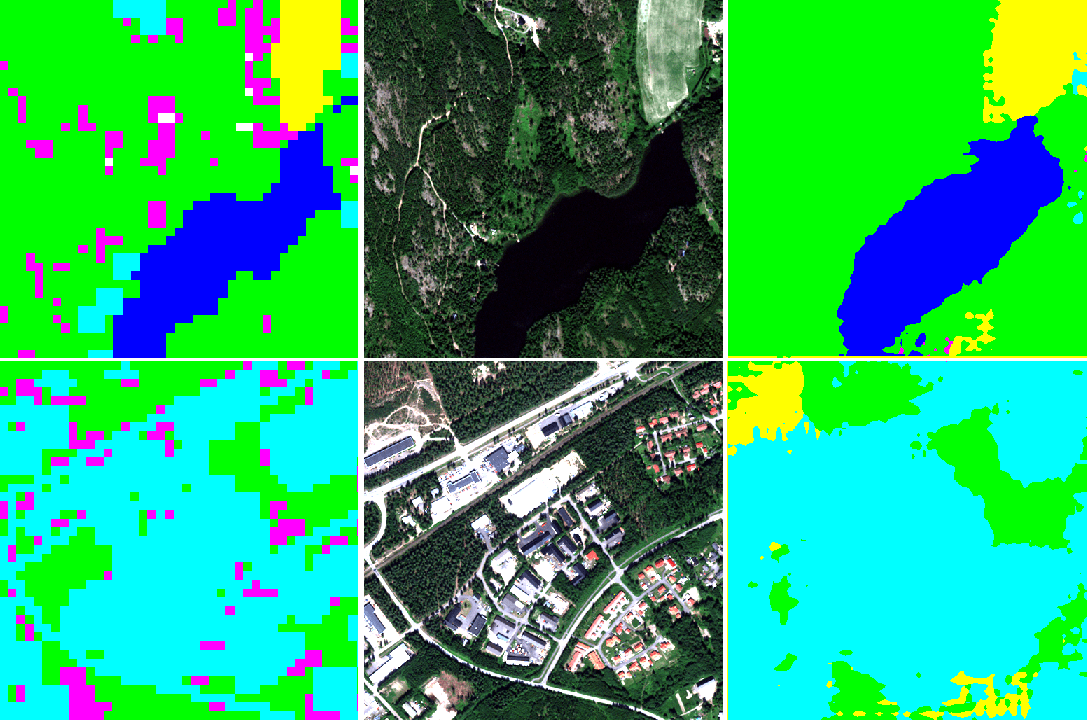}
            \end{center}
            \caption{Results of training with domain adaptation from WorldView-2 (Germany) to WorldView-2 (Finland). \textit{Right:} output of model. \textit{middle:} test images from WorldView-2 (Finland) dataset. \textit{Left:} ground truth images from WorldView-2 (Finland) dataset.}
            \label{fig:WVGR2WVFIcycle}
    \end{figure}

The results obtained from \textit{WV2GR to WV2FI} using domain adaptation have improved significantly compared to the previous case. As seen in Table~\ref{wv2wv}, the MIoU score rose to 27. Fig~\ref{fig:WVGR2WVFIcycle} displays a sample from this experiment. Although the results are not very precise, especially on small details, it is a distinct improvement compared to the results without domain adaption (see Fig~\ref{fig:WVGR2WVFInocycle}).
 

\subsubsection{Deeplabv3+ results}
    
In addition to the previous experiments, we tested Deeplabv3+ as the segmentation network. Replacing the backbone with Deeplabv3+ did not bring significant improvements compared to Deeplabv2 as shown in Table~\ref{wv2dg}, Table~\ref{sen2dg}, Table~\ref{sen2wv}, Table~\ref{wv2pl}, and Table~\ref{wv2wv}. The reason behind that is the lack of accurate labels, which might be bottlenecking the performance. In fact, the results diverge after a few epochs because the network becomes too good at mimicking the errors in the GT images. So in order to make use of the full potential of deep learning methods it is important to have a very good and precise data in land cover mapping.

\FloatBarrier
\section{Conclusion}
In this paper, we addressed three problems relating to land use and land cover mapping: lack of labeled datasets, inaccessibility to some satellite imagery, and the difference in the available spectral bands. The approach we adopted is using domain adaptation on datasets we have built from RGB imagery of different satellites covering different areas. The experimental results show that domain adaption improved the mapping considerably even with the usage of RGB images only. The results extend to images from areas with considerably different land cover types. 
Our findings suggest the possibility to further improve the results by designing a more specialized model for satellite imagery.

\section*{Acknowledgment}
This research was financially supported by Business Finland (Grant no. 1259/31/2018). Authors would also like to thank the project partners from VTT Technical Research Centre of Finland for providing their datasets and technical support.

\bibliographystyle{IEEEtran}
\bibliography{IEEEtran}

\begin{thebibliography}{10}
\providecommand{\url}[1]{#1}
\csname url@samestyle\endcsname
\providecommand{\newblock}{\relax}
\providecommand{\bibinfo}[2]{#2}
\providecommand{\BIBentrySTDinterwordspacing}{\spaceskip=0pt\relax}
\providecommand{\BIBentryALTinterwordstretchfactor}{4}
\providecommand{\BIBentryALTinterwordspacing}{\spaceskip=\fontdimen2\font plus
\BIBentryALTinterwordstretchfactor\fontdimen3\font minus
  \fontdimen4\font\relax}
\providecommand{\BIBforeignlanguage}[2]{{%
\expandafter\ifx\csname l@#1\endcsname\relax
\typeout{** WARNING: IEEEtran.bst: No hyphenation pattern has been}%
\typeout{** loaded for the language `#1'. Using the pattern for}%
\typeout{** the default language instead.}%
\else
\language=\csname l@#1\endcsname
\fi
#2}}
\providecommand{\BIBdecl}{\relax}
\BIBdecl

\bibitem{copernicus_open}
\BIBentryALTinterwordspacing
``Copernicus open access hub.'' [Online]. Available:
  \url{https://scihub.copernicus.eu/dhus/}
\BIBentrySTDinterwordspacing

\bibitem{usgs}
\BIBentryALTinterwordspacing
``Usgs.'' [Online]. Available: \url{https://earthexplorer.usgs.gov/}
\BIBentrySTDinterwordspacing

\bibitem{doi:10.1080/01431169208904145}
\BIBentryALTinterwordspacing
W.~AHMAD, L.~B. JUPP, and M.~NUNEZ, ``Land cover mapping in a rugged terrain
  area using landsat mss data,'' \emph{International Journal of Remote
  Sensing}, vol.~13, no.~4, pp. 673--683, 1992. [Online]. Available:
  \url{https://doi.org/10.1080/01431169208904145}
\BIBentrySTDinterwordspacing

\bibitem{corinepdf}
\BIBentryALTinterwordspacing
B.~Kosztra, G.~Büttner, G.~Hazeu, and S.~Arnold, ``Updated clc illustrated
  nomenclature guidelines.'' [Online]. Available:
  \url{https://land.copernicus.eu/user-corner/technical-library/corine-land-cover-nomenclature-guidelines/docs/pdf/CLC2018\_Nomenclature\_illustrated\_guide\_20170930\\.pdf}
\BIBentrySTDinterwordspacing

\bibitem{MODIS}
\BIBentryALTinterwordspacing
``Modis web.'' [Online]. Available:
  \url{https://modis.gsfc.nasa.gov/data/dataprod/mod12.php}
\BIBentrySTDinterwordspacing

\bibitem{7301382}
O.~A.~B. {Penatti}, K.~{Nogueira}, and J.~A. {dos Santos}, ``Do deep features
  generalize from everyday objects to remote sensing and aerial scenes
  domains?'' in \emph{2015 IEEE Conference on Computer Vision and Pattern
  Recognition Workshops (CVPRW)}, June 2015, pp. 44--51.

\bibitem{DBLP:journals/corr/abs-1805-06561}
\BIBentryALTinterwordspacing
I.~Demir, K.~Koperski, D.~Lindenbaum, G.~Pang, J.~Huang, S.~Basu, F.~Hughes,
  D.~Tuia, and R.~Raskar, ``Deepglobe 2018: {A} challenge to parse the earth
  through satellite images,'' \emph{CoRR}, vol. abs/1805.06561, 2018. [Online].
  Available: \url{http://arxiv.org/abs/1805.06561}
\BIBentrySTDinterwordspacing

\bibitem{deepglobe.densenet}
C.~Tian, C.~Li, and J.~Shi, ``Dense fusion classmate network for land cover
  classification,'' 06 2018, pp. 262--2624.

\bibitem{DBLP:journals/corr/HuangLW16a}
\BIBentryALTinterwordspacing
G.~Huang, Z.~Liu, and K.~Q. Weinberger, ``Densely connected convolutional
  networks,'' \emph{CoRR}, vol. abs/1608.06993, 2016. [Online]. Available:
  \url{http://arxiv.org/abs/1608.06993}
\BIBentrySTDinterwordspacing

\bibitem{Kuo2018DeepAN}
T.-S. Kuo, K.-S. Tseng, J.~Yan, Y.-C. Liu, and Y.-C.~F. Wang, ``Deep
  aggregation net for land cover classification,'' \emph{2018 IEEE/CVF
  Conference on Computer Vision and Pattern Recognition Workshops (CVPRW)}, pp.
  247--2474, 2018.

\bibitem{syke}
\BIBentryALTinterwordspacing
``Metatietopalvelu.'' [Online]. Available:
  \url{http://metatieto.ymparisto.fi:8080/geoportal/catalog/search/resource/
  details.page?uuid=\%7B6833C06E-BF77-4F0B-A066-B94AE98392EA\%7D}
\BIBentrySTDinterwordspacing

\bibitem{CLC_GR}
\BIBentryALTinterwordspacing
``Bkg - corine land cover.'' [Online]. Available:
  \url{https://www.bkg.bund.de/DE/Ueber-das-BKG/Geoinformation/Fernerkundung/Landbedeckungsmodell/
  CorineLandCover/clc.html}
\BIBentrySTDinterwordspacing

\bibitem{1455595}
A.~{Rosenfeld} and L.~S. {Davis}, ``Image segmentation and image models,''
  \emph{Proceedings of the IEEE}, vol.~67, no.~5, pp. 764--772, May 1979.

\bibitem{ecognition}
\BIBentryALTinterwordspacing
``What is ecognition?'' [Online]. Available:
  \url{https://geospatial.trimble.com/what-is-ecognition}
\BIBentrySTDinterwordspacing

\bibitem{doi:10.1080/10106049.2013.768300}
\BIBentryALTinterwordspacing
M.~S. Tehrany, B.~Pradhan, and M.~N. Jebuv, ``A comparative assessment between
  object and pixel-based classification approaches for land use/land cover
  mapping using spot 5 imagery,'' \emph{Geocarto International}, vol.~29,
  no.~4, pp. 351--369, 2014. [Online]. Available:
  \url{https://doi.org/10.1080/10106049.2013.768300}
\BIBentrySTDinterwordspacing

\bibitem{KHATAMI201689}
\BIBentryALTinterwordspacing
R.~Khatami, G.~Mountrakis, and S.~V. Stehman, ``A meta-analysis of remote
  sensing research on supervised pixel-based land-cover image classification
  processes: General guidelines for practitioners and future research,''
  \emph{Remote Sensing of Environment}, vol. 177, pp. 89 -- 100, 2016.
  [Online]. Available:
  \url{http://www.sciencedirect.com/science/article/pii/S0034425716300578}
\BIBentrySTDinterwordspacing

\bibitem{Deeplabv3_plus}
\BIBentryALTinterwordspacing
L.~Chen, Y.~Zhu, G.~Papandreou, F.~Schroff, and H.~Adam, ``Encoder-decoder with
  atrous separable convolution for semantic image segmentation,'' \emph{CoRR},
  vol. abs/1802.02611, 2018. [Online]. Available:
  \url{http://arxiv.org/abs/1802.02611}
\BIBentrySTDinterwordspacing

\bibitem{rs10060973}
\BIBentryALTinterwordspacing
H.~A. Arief, G.-H. Strand, H.~Tveite, and U.~G. Indahl, ``Land cover
  segmentation of airborne lidar data using stochastic atrous network,''
  \emph{Remote Sensing}, vol.~10, no.~6, 2018. [Online]. Available:
  \url{http://www.mdpi.com/2072-4292/10/6/973}
\BIBentrySTDinterwordspacing

\bibitem{DBLP:journals/corr/abs-1802-03601}
\BIBentryALTinterwordspacing
M.~Wang and W.~Deng, ``Deep visual domain adaptation: {A} survey,''
  \emph{CoRR}, vol. abs/1802.03601, 2018. [Online]. Available:
  \url{http://arxiv.org/abs/1802.03601}
\BIBentrySTDinterwordspacing

\bibitem{DBLP:journals/corr/abs-1812-02849}
\BIBentryALTinterwordspacing
G.~Wilson and D.~J. Cook, ``Adversarial transfer learning,'' \emph{CoRR}, vol.
  abs/1812.02849, 2018. [Online]. Available:
  \url{http://arxiv.org/abs/1812.02849}
\BIBentrySTDinterwordspacing

\bibitem{gan}
I.~Goodfellow, J.~Pouget-Abadie, M.~Mirza, B.~Xu, D.~Warde-Farley, S.~Ozair,
  A.~Courville, and Y.~Bengio, ``Generative adversarial nets,'' in
  \emph{Advances in neural information processing systems}, 2014, pp.
  2672--2680.

\bibitem{adda}
\BIBentryALTinterwordspacing
E.~Tzeng, J.~Hoffman, K.~Saenko, and T.~Darrell, ``Adversarial discriminative
  domain adaptation,'' \emph{CoRR}, vol. abs/1702.05464, 2017. [Online].
  Available: \url{http://arxiv.org/abs/1702.05464}
\BIBentrySTDinterwordspacing

\bibitem{cycle}
\BIBentryALTinterwordspacing
J.~Zhu, T.~Park, P.~Isola, and A.~A. Efros, ``Unpaired image-to-image
  translation using cycle-consistent adversarial networks,'' \emph{CoRR}, vol.
  abs/1703.10593, 2017. [Online]. Available:
  \url{http://arxiv.org/abs/1703.10593}
\BIBentrySTDinterwordspacing

\bibitem{Zhang2017CurriculumDA}
Y.~Zhang, P.~David, and B.~Gong, ``Curriculum domain adaptation for semantic
  segmentation of urban scenes,'' \emph{2017 IEEE International Conference on
  Computer Vision (ICCV)}, pp. 2039--2049, 2017.

\bibitem{Hoffman2016FCNsIT}
J.~Hoffman, D.~Wang, F.~Yu, and T.~Darrell, ``Fcns in the wild: Pixel-level
  adversarial and constraint-based adaptation,'' \emph{ArXiv}, vol.
  abs/1612.02649, 2016.

\bibitem{No_More_Discrimination}
Y.-H. Chen, W.-Y. Chen, Y.-T. Chen, B.-C. Tsai, Y.-C.~F. Wang, and M.~Sun, ``No
  more discrimination: Cross city adaptation of road scene segmenters,'' 10
  2017, pp. 2011--2020.

\bibitem{Hong_2018_CVPR}
W.~Hong, Z.~Wang, M.~Yang, and J.~Yuan, ``Conditional generative adversarial
  network for structured domain adaptation,'' in \emph{The IEEE Conference on
  Computer Vision and Pattern Recognition (CVPR)}, June 2018.

\bibitem{DBLP:journals/corr/abs-1903-12212}
\BIBentryALTinterwordspacing
W.~Chang, H.~Wang, W.~Peng, and W.~Chiu, ``All about structure: Adapting
  structural information across domains for boosting semantic segmentation,''
  \emph{CoRR}, vol. abs/1903.12212, 2019. [Online]. Available:
  \url{http://arxiv.org/abs/1903.12212}
\BIBentrySTDinterwordspacing

\bibitem{DBLP:journals/corr/abs-1802-10349}
\BIBentryALTinterwordspacing
Y.~Tsai, W.~Hung, S.~Schulter, K.~Sohn, M.~Yang, and M.~Chandraker, ``Learning
  to adapt structured output space for semantic segmentation,'' \emph{CoRR},
  vol. abs/1802.10349, 2018. [Online]. Available:
  \url{http://arxiv.org/abs/1802.10349}
\BIBentrySTDinterwordspacing

\bibitem{Vu_2019_CVPR}
T.-H. Vu, H.~Jain, M.~Bucher, M.~Cord, and P.~Perez, ``Advent: Adversarial
  entropy minimization for domain adaptation in semantic segmentation,'' in
  \emph{The IEEE Conference on Computer Vision and Pattern Recognition (CVPR)},
  June 2019.

\bibitem{Cordts2016Cityscapes}
M.~Cordts, M.~Omran, S.~Ramos, T.~Rehfeld, M.~Enzweiler, R.~Benenson,
  U.~Franke, S.~Roth, and B.~Schiele, ``The cityscapes dataset for semantic
  urban scene understanding,'' in \emph{Proc. of the IEEE Conference on
  Computer Vision and Pattern Recognition (CVPR)}, 2016.

\bibitem{Richter_2016_ECCV}
S.~R. Richter, V.~Vineet, S.~Roth, and V.~Koltun, ``Playing for data: {G}round
  truth from computer games,'' in \emph{European Conference on Computer Vision
  (ECCV)}, ser. LNCS, B.~Leibe, J.~Matas, N.~Sebe, and M.~Welling, Eds., vol.
  9906.\hskip 1em plus 0.5em minus 0.4em\relax Springer International
  Publishing, 2016, pp. 102--118.

\bibitem{Ros_2016_CVPR}
G.~Ros, L.~Sellart, J.~Materzynska, D.~Vazquez, and A.~M. Lopez, ``The synthia
  dataset: A large collection of synthetic images for semantic segmentation of
  urban scenes,'' in \emph{The IEEE Conference on Computer Vision and Pattern
  Recognition (CVPR)}, June 2016.

\bibitem{Chang_2019_CVPR}
W.-L. Chang, H.-P. Wang, W.-H. Peng, and W.-C. Chiu, ``All about structure:
  Adapting structural information across domains for boosting semantic
  segmentation,'' in \emph{The IEEE Conference on Computer Vision and Pattern
  Recognition (CVPR)}, June 2019.

\bibitem{Li_2019_CVPR}
Y.~Li, L.~Yuan, and N.~Vasconcelos, ``Bidirectional learning for domain
  adaptation of semantic segmentation,'' in \emph{The IEEE Conference on
  Computer Vision and Pattern Recognition (CVPR)}, June 2019.

\bibitem{Vivid}
\BIBentryALTinterwordspacing
Digitalglobe basemap +vivid. [Online]. Available:
  \url{https://dg-cms-uploads-production.s3.amazonaws.com/uploads/document/file/2/\\DG\_Basemap\_Vivid\_DS\_1.pdf}
\BIBentrySTDinterwordspacing

\bibitem{esatpm}
\BIBentryALTinterwordspacing
``Worldview-2 european cities - view data product - earth online - esa.''
  [Online]. Available:
  \url{https://earth.esa.int/web/guest/-/worldview-2-european-cities-dataset}
\BIBentrySTDinterwordspacing

\bibitem{8898776}
S.~{Mohajerani} and P.~{Saeedi}, ``Cloud-net: An end-to-end cloud detection
  algorithm for landsat 8 imagery,'' in \emph{IGARSS 2019 - 2019 IEEE
  International Geoscience and Remote Sensing Symposium}, July 2019, pp.
  1029--1032.

\bibitem{v100}
\BIBentryALTinterwordspacing
``Nvidia tesla v100 data center gpu.'' [Online]. Available:
  \url{https://www.nvidia.com/en-us/data-center/tesla-v100/}
\BIBentrySTDinterwordspacing

\bibitem{Tian_2018_CVPR_Workshops}
C.~Tian, C.~Li, and J.~Shi, ``Dense fusion classmate network for land cover
  classification,'' in \emph{The IEEE Conference on Computer Vision and Pattern
  Recognition (CVPR) Workshops}, June 2018.

\end{thebibliography}

\vspace{12pt}
\end{document}